\documentclass[10pt,twocolumn,letterpaper]{article}

\usepackage{cvpr}              % To produce the CAMERA-READY version
\definecolor{cvprblue}{rgb}{0.21,0.49,0.74}
\usepackage[pagebackref,breaklinks,colorlinks,allcolors=cvprblue]{hyperref}
\usepackage{pifont}
\usepackage{multirow}
\usepackage{xcolor}
\usepackage{makecell}
\usepackage{caption}
\usepackage{amsmath}
\usepackage{bm}
\usepackage[table]{xcolor}

\def\paperID{***} % *** Enter the Paper ID here
\def\confName{CVPR}
\def\confYear{2026}

\title{Narrative Weaver:  Towards Controllable Long-Range Visual Consistency with Multi-Modal Conditioning}
\author{
Zhengjian Yao$^{1,}$\thanks{Work done during an internship at Kuaishou Technology} \quad
Yongzhi Li$^{2,}$\thanks{Project Leader, $^\ddagger$Corresponding Authors.} \quad
Xinyuan Gao$^{2}$ \quad
Quan Chen$^{2,\ddagger}$  \quad
Peng Jiang$^{2}$ \quad
Yanye Lu$^{1,\ddagger}$ \\
$^{1}$Peking University \quad
$^{2}$Kuaishou Technology \\
{\tt\small zj.yao@stu.pku.edu.cn\quad yanye.lu@pku.edu.cn} \\
{\tt\small \{liyongzhi03, gaoxinyuan, chenquan06, jiangpeng\}@kuaishou.com} 
}

\begin{document}
\maketitle

\begin{abstract}
We present Narrative Weaver, a novel framework that addresses a fundamental challenge in generative AI: achieving multi-modal controllable, long-range, and consistent visual content generation. While existing models excel at generating high-fidelity short-form visual content, they struggle to maintain narrative coherence and visual consistency across extended sequences—a critical limitation for real-world applications such as filmmaking and e-commerce advertising.
Narrative Weaver introduces the first holistic solution that seamlessly integrates three essential capabilities: fine-grained control, automatic narrative planning, and long-range coherence. Our architecture combines a Multimodal Large Language Model (MLLM) for high-level narrative planning with a novel fine-grained control module featuring a dynamic Memory Bank that prevents visual drift.
To enable practical deployment, we develop a progressive, multi-stage training strategy that efficiently leverages existing pre-trained models, achieving state-of-the-art performance even with limited training data. Recognizing the absence of suitable evaluation benchmarks, we construct and release the E-commerce Advertising Video Storyboard Dataset (EAVSD)—the first comprehensive dataset for this task, containing over 330K high-quality images with rich narrative annotations.
Through extensive experiments across three distinct scenarios (controllable multi-scene generation, autonomous storytelling, and e-commerce advertising), we demonstrate our method’s superiority while opening new possibilities for AI-driven content creation. 
\end{abstract}    
\section{Introduction}
\label{sec:intro}
Generative Artificial Intelligence, propelled by advances in diffusion models~\cite{ho2020ddpm,song2019scorematching}, has revolutionized visual content creation. Pioneering systems such as Sora~\cite{sora2025}, Veo~\cite{wiedemer2025video}, and Midjourney~\cite{midjourney} exhibit remarkable capabilities in producing high-fidelity images and videos. The open-source community is keeping pace with notable releases like Wan~\cite{wan2025wan}, CogVideo~\cite{hong2022cogvideo}, Qwen-Image~\cite{wu2025qwenimage}, and Flux~\cite{flux1dev2024}.
Despite these strides, a critical challenge remains unaddressed: the automatic planning and generation of long-range visual narratives with strict semantic and visual consistency.

%—————————————————————————————————————————————————— V2 ——————————————————————————————————————————————————————————————
% Generative Artificial Intelligence, particularly driven by Diffusion models~\cite{ho2020ddpm, song2019scorematching}, has achieved remarkable success in visual content creation. Landmark models like Sora~\cite{sora2025}, Veo~\cite{wiedemer2025video}, and Midjourney~\cite{midjourney} have demonstrated astounding capabilities in generating high-fidelity videos and images. The open-source community is also rapidly advancing with notable efforts such as Wan~\cite{wan2025wan}, CogVideo~\cite{hong2022cogvideo}, Qwen-Image~\cite{wu2025qwenimage}, Flux~\cite{flux1dev2024}. However, despite this rapid progress, a fundamental and largely unexplored research gap persists: the automatic planning and generation of long-range visual content that maintains strict semantic and visual consistency.
%——————————————————————————————————————————————————————————————————————————————————————————————————————————————————————

%—————————————————————————————————————————————————— V2 ——————————————————————————————————————————————————————————————
This limitation severely hampers real-world applications that demand narrative continuity. 
In video, even top models fail beyond short clips, struggling to maintain the consistent characters, backgrounds, and storylines essential for effective storytelling or advertising~\cite{sora2025,wiedemer2025video,wan2025wan}. A parallel challenge exists for static images, where powerful tools are confined to single-frame operations~\cite{batifol2025kontext, wu2025qwenimage, flux1dev2024}. 
Although some works have attempted to integrate planning, their reliance on purely textual conditioning renders them incapable of delivering the controllable visual output required in practical scenarios~\cite{zhuang2024vlogger, xiao2025videoauteur, long2024videostudio, lin2023videodirectorgpt, yan2025longtake, wu2025mint, hu2025mvar, hu2026geometry}. 
These observations highlight the problem: the absence of a unified framework that synergizes narrative planning with fine-grained, visually-grounded control for long-range coherence.

%——————————————————————————————————————————————————————————————————————————————————————————————————————————————————————
% This gap creates significant limitations for practical applications. In video generation, SOTA models [3, 4] can produce impressive, smooth clips, yet they are typically confined to short durations (e.g., 5-10 seconds). This is insufficient for scenarios demanding narrative continuity, such as filmmaking, or for commercial applications like marketing videos, where a specific product must remain visually identical across disparate scenes and actions. Existing methods simply fail to maintain this long-range temporal and object consistency.

% In the static image domain, parallel challenges exist. While fine-grained image editing and generation have achieved high fidelity [8, 9], these methods typically operate on single images and rely on meticulous, manual prompt engineering. They lack the native capability for automated long-form narrative planning—for example, generating a coherent multi-panel comic strip from a single prompt. While some works have attempted to integrate planning with generation [10, 11], they are often restricted to text-only inputs. This critical limitation makes it impossible to enforce precise visual fidelity to an image condition. For marketing, where a specific product SKU, brand logo, or visual identity is non-negotiable, text-only planners are fundamentally inadequate.

To address this challenge, we present Narrative Weaver, a framework that unifies multimodal-conditioned narrative planning with fine-grained control for long-range visual consistency.
Narrative Weaver first employs a MLLM as a "director," which takes initial visual and textual context to devise a high-level storyboard. This storyboard is then translated into explicit semantic concepts and spatial layouts via a learnable query module. 
% Critically, a dynamic memory bank mitigates visual drift by anchoring each generative step to initial visual conditions and prior frames, thus ensuring long-range coherence. 
To ensure long-range coherence, a dynamic memory bank mitigates visual drift by anchoring each generative step to initial visual conditions and prior frames. 
Furthermore, we introduce a multi-stage training strategy that enables our model to achieve leading performance in a data-efficient manner.

Realizing and rigorously evaluating such a system is obstructed by a critical data scarcity: no existing dataset~\cite{ju2025ci-vid, yang2025seed-story, wu2025omnigen2} provides the necessary multi-modal conditioning format of ($\texttt{text}, \texttt{image}$) $\mapsto$ ($ \texttt{text}, \{\texttt{Image}_i\}_{i=1}^N $).
To bridge this void, we construct the E-commerce Advertising Video Storyboard Dataset (EAVSD). This dataset is specifically curated for e-commerce marketing, where unwavering visual consistency is a commercial necessity for maintaining brand identity. 
It provides triplets of (product image, description, marketing goal) meticulously mapped to multi-scene storyboards.
We rigorously evaluate our framework through extensive experiments on existing benchmarks and our new EAVSD. 
The results affirm the novelty of our approach and its superiority over previous methods.
\section{Related Works}
\label{sec:related_works}

\textbf{Extending Visual Duration.} Research on extending visual duration broadly follows three directions. The first focuses on context compression~\cite{wang2025lingen, jin2024pyramidal, lin2024opensora}, targeting either the attention mechanism or token sequence length. For instance, FramePack~\cite{zhang2025framepack} compresses historical information into a fixed-length sequence, MoC~\cite{cai2025moc} adaptively attends to critical historical tokens, and LTX-Video~\cite{hacohen2024ltx} employs a VAE for higher compression rates. The second category adopts a chunk-based strategy~\cite{qi2025mask2dit, huang2024gdt, cai2024ditctrl, yan2025longtake, atzmon2024videostoryboarding}, generating long videos segment by segment. Methods like TokensGen~\cite{ouyang2025tokensgen} and AnimeShooter~\cite{qiu2025animeshooter} use techniques like learnable queries or FIFO-diffusion~\cite{kim2024fifo} to encode preceding chunks into compact representations, while others employ more direct auto-regressive frameworks to condition subsequent generation on prior segments~\cite{chen2024diffusionforcing, song2025history, huang2025selfforcing, cui2025selfforcing++, li2025svi, gu2025far, huang2024owl1omniworldmodel}. The third direction utilizes keyframe-based generation~\cite{meng2025holocine, zheng2024vgot, long2024videostudio, zhao2024moviedreamer, xiao2025videoauteur, mao2024story_adapter}, an efficient approach that preserves the capabilities of pre-trained models. StoryDiffusion~\cite{zhou2024storydiffusion}, for example, generates keyframes from sub-prompts, and CaptainCinema~\cite{xiao2025captain} integrates past keyframes via a GoldenMem module to contextualize subsequent ones.
% However, their conditioning is limited to text or previous frames, failing to ground the entire extended sequence in an initial, user-provided visual anchor. Our work, in contrast, is not merely about duration but about controllable, narratively coherent extension from multi-modal inputs.
Their conditioning is limited to text or previous frames, failing to ground the extended sequence in an initial visual anchor. Our work focuses on controllable, narratively coherent extension from multi-modal inputs, not merely duration.

\textbf{Narrative Visual Generation.} Advances in large language models (LLMs) and unified architectures~\cite{chen2025blip3, deng2025bagel, wu2025omnigen2, xie2025showo} have enabled the generation of multi-modal content. However, most existing methods are limited to single-round generation and lack the capacity for sophisticated interleaved reasoning and planning. Crucially, they often fail to maintain visual coherence across generated content. Diverging strategies have emerged to address this: VideoAuteur~\cite{xiao2025videoauteur} introduces an interleaved VLM director, whereas LCT~\cite{guo2025lct} fine-tunes MM-DiTs directly. Recent methods—including VideoDirectorGPT~\cite{lin2023videodirectorgpt}, Vlogger~\cite{zhuang2024vlogger}, Animate-a-Story~\cite{he2023animateastory}, IC-LoRA~\cite{huang2024iclora}, and StoryDiffusion~\cite{zhou2024storydiffusion}—have demonstrated improved ability to produce visually coherent sequences from textual narratives. While most prior work~\cite{wu2025movieagent, ren2025videorag} focuses on generating semantically consistent image sets, our approach enables autonomous narrative generation without complex pipelines, achieving a more streamlined and efficient implementation.
%————————————————————————————————————————————————————————————————————————————————————————————————————————————————————————————

\textbf{Datasets for Consistency Visual Generation.} Existing large-scale video generation datasets such as Koala-36M~\cite{wang2025koala}, Panda-70M~\cite{chen2024panda}, HD-VG-130M~\cite{wang2025hdvg}, and MiraData~\cite{ju2024miradata} provide valuable resources for training generative models, yet they mainly consist of short clips lasting only a few seconds, limiting their ability to model extended storylines. Effective long-range visual generation requires datasets supporting conditional image-to-multiframe generation with coherent narrative structures. 
% However, current resources remain insufficient: large-scale corpora like OmniGen2~\cite{wu2025omnigen2} lack conditional image grounding, annotated video sets such as CI-VID~\cite{ju2025ci-vid} contain only a few clips per instance, and narrative-oriented datasets like StoryStream~\cite{yang2025seed-story} are restricted to simplified animation scenarios. These limitations highlight the need for datasets that jointly address visual grounding, narrative planning, and multi-frame consistency.
However, Current resources remain limited: large-scale corpora like OmniGen2~\cite{wu2025omnigen2} lack conditional image grounding, video sets such as CI-VID~\cite{ju2025ci-vid} contain few clips per instance, and narrative datasets like StoryStream~\cite{yang2025seed-story} cover only simplified animations. These gaps highlight the need for datasets addressing visual grounding, narrative planning, and multi-frame consistency.
% This scarcity of large-scale, narratively-grounded video data presents a fundamental barrier to developing models that maintain long-term coherence. To bridge this gap, we open-source a curated e-commerce long-video keyframe dataset to facilitate future research.

%-------------------------------------------------------------------------
% \subsection{Footnotes}

% Please use the footnotes\footnote{This is what a footnote looks like.
% It often distracts the reader from the main flow of the argument.} sparingly.
% Indeed, try to avoid footnotes altogether and include necessary peripheral observations in the text (within parentheses, if you prefer, as in this sentence).
% If you wish to use a footnote, place it at the bottom of the column on the page on which it is referenced.
% Use Times 8-point type, single-spaced.

%-------------------------------------------------------------------------

%-------------------------------------------------------------------------
\begin{figure*}
  \centering
    \includegraphics[width=.98\linewidth]{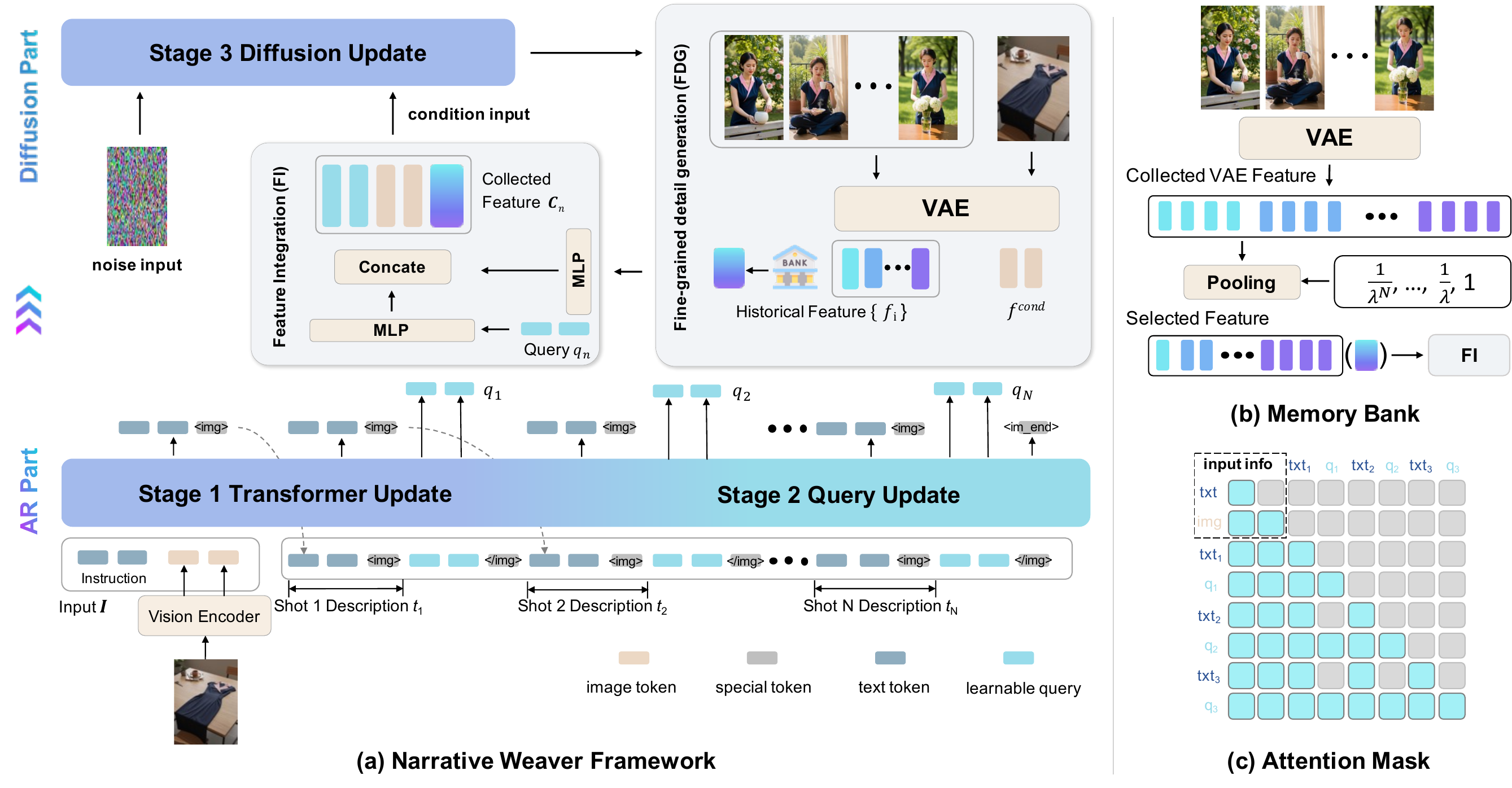}
    \vspace{-3mm}
  \caption{\textbf{Narrative Weaver Overview.} (a) Narrative Weaver Framework: This system utilizes a hybrid design that integrates Autoregressive (AR) and Diffusion models. The bottom panel illustrates a Multimodal Large Language Model (MLLM) acting as the AR model, responsible for generating narrative plans in textual form and encoding historical information into learnable queries. During the diffusion generation stage, a dynamic memory bank encodes initial conditions and prior outputs to prevent visual content drift. (b) Memory Bank: We employ a series-based decay of prior visual feature length to ensure a bounded total memory length. (c) Attention Mask: A specially designed Attention Mask ensures efficient training, where gray areas are ignored during processing.}
  \label{fig:method}
  \vspace{-5mm}
\end{figure*}

\section{Methods}
\label{sec:methods}
% This section delineates the architecture and training methodology of Narrative Weaver. Specifically, \cref{sec:model} introduces our core model architecture, explaining its inherent mechanisms for consistency preservation and autonomous narrative generation. \cref{sec:training} elaborates on the multi-stage training pipeline, detailing how we cultivate competitive keyframe generation potential with limited data. Finally, \cref{sec:data} outlines our data curation methodology, with a focus on constructing a high-quality keyframe dataset from e-commerce sources.

\subsection{Narrative Weaver Framework}
\label{sec:model}

\textbf{Framework Overview.} We present Narrative Weaver, a hybrid Autoregressive (AR) + Diffusion framework~\cite{chen2025blip3,ge2024seedx,pan2025metaquery,hu2025omni-view}. As illustrated in \cref{fig:method} (a), the AR part encodes historical context, while the diffusion model decodes this representation to generate coherent visual content. 
Upon receiving the input $\bm{I}$, which comprises condition images and user instructions, the AR part performs two key functions: explicitly planning future narrative logic $\bm{T} = \{t_i\}_{i=0,1,\dots}$ in textual form, and condensing historical multimodal information into a compact high-level, learnable query representation $\bm{Q} = \{q_i\}_{i=0,1,\dots}$.
Next, fine-grained VAE-encoded features $f^{\text{cond}}$ from input conditioning image is integrated with $\bm{Q}$, forming a comprehensive conditioning signal $\bm{C} = \{c_i = [q_i; f^{\text{cond}}]\}_{i=0,1,\dots}$. This fundamental fused conditioning set $\bm{C}$ is then fed into the diffusion model to generate highly coherent visual sequences.
% For simplicity, we use keyframe as a representative of visual content in the following text.

\textbf{Multimodal Interaction.} 
The AR component of Narrative Weaver is designed to simultaneously perform narrative planning ($\bm{T}$) and high-level visual content aggregation ($\bm{Q}$). To balance effective multi-modal information exchange and efficient utilization of pre-trained MLLM prior knowledge, while preventing the newly introduced learnable queries from disturbing the original model outputs, we propose a dynamic causal attention mask (\cref{fig:method} (c)).

Within this configuration, each learnable query $q_n$ is granted comprehensive access to the full multimodal context, encompassing the input $\bm{I}$, all narrative text tokens up to the current step $\{t_j\}_{j=0}^{n}$, and previously aggregated queries $\{q_k\}_{k=0}^{n-1}$. This extensive conditioning ensures that the subsequent keyframe generation remains visually coherent and adheres strictly to the evolving narrative guidance. In contrast, textual tokens are constrained by a causal attention mechanism, attending only to preceding text tokens. This design choice facilitates robust narrative planning through standard next-token prediction, where each $t_n$ is generated based on the conditional probability:
\begin{equation}
t_n \sim P(t_n | \bm{I}, \{t_j\}_{j=0}^{n-1}).
\label{eq:eq1}
\end{equation}

Furthermore, during training, special tokens \texttt{<img>} and \texttt{</img>} are introduced to bracket the learnable query sequence. Through this mechanism, the model learns to: 1) predict the appropriate timing for visual output, and 2) subsequently continue with textual content planning after generating an image. 
This holistic framework enables the model to acquire textual planning capabilities with minimal data ($\sim$  5K), while simultaneously achieving high-level consistency across the generated visual content.

% \begin{figure}[!ht]
%   \centering
%    \includegraphics[width=0.7\linewidth]{sec/Images/AttentionMask.pdf}
%     \vspace{-2mm}
%    \caption{Attention Mask of Narrative Weaver.}
%    \label{fig:attention_mask}
% \end{figure}

\textbf{Memory Bank.} To ensure temporal stability across sequentially generated images, which is crucial for downstream applications such as comics and film generation, we propose a memory bank designed to encode features from preceding images (\cref{fig:method} (b)).

Specifically, the memory bank caches VAE features of already generated images, denoted as $\bm{B} = \{f_i\}_{i=1,2,\dots}$, where $f_i \in \mathbb{R}^{l \times d}$ represents the VAE feature of the $i$-th generated image, and $l$ is its feature length. 
% As in~\cite{zhang2025framepack, xiao2025captain}, when generating the $n$-th image, we incorporate features from $T$ preceding generated images, $\{f_{n-k}\}_{k=1}^T$.
Following~\cite{zhang2025framepack, xiao2025captain}, we incorporate features from $T$ preceding generated images, $\{f_{n-k}\}_{k=1}^T$, when generating the $n$-th image.
To manage computational cost and emphasize recent history, we apply an average pooling operation to each $f_{n-k}$ to obtain $\hat{f}_{n-k}$. 
This pooling reduces the feature length by a decay factor $\lambda > 1$, such that the length of $\hat{f}_{n-k}$ becomes $l / \lambda^{k-1}$. This geometrically decaying length ensures a bounded total sequence length for the aggregated memory features:
\begin{align}
L & = \sum_{k=1}^{T} \text{len}(\hat{f}_{n-k}) = \sum_{j=0}^{T-1} \frac{l}{\lambda^j} \nonumber \\
&= l \frac{1 - (1/\lambda)^T}{1 - 1/\lambda} < l \frac{\lambda}{\lambda - 1} \quad (\text{for } \lambda > 1).
\end{align}

Finally, the comprehensive conditioning signal $\mathbf{C}_n$ for generating the $n$-th keyframe is formed by concatenating the current learnable query $q_n$, the VAE feature of the current input conditioning image $f^{\text{cond}}$, and the pooled features from the memory bank:
\begin{equation}
\mathbf{C}_n = \text{Concat}(q_n, f^{\text{cond}}, \hat{f}_{n-1}, \dots, \hat{f}_{n-T}),
\end{equation}
where $T$ is a hyperparameter controlling the number of preceding images considered.

% \textbf{Efficiency Analysis.} By employing the MLLM as a mediator and learnable queries to encode inter-image coherence, our framework reduces the DiT's computational complexity from quadratic to linear growth with respect to the number of images~\cite{huang2024iclora, huang2024gdt,cai2025moc,dalal2025ttt,xiao2025captain}. Furthermore, this design shifts the computational bottleneck to the MLLM component, which benefits from the abundant existing optimization techniques~\cite{dao2022flashattention, pytorch_flexattention, unsloth_ai}, thus decreasing overall resource consumption while maintaining generation quality. At inference time, this design enables parallel narrative planning and visual generation, offering improved efficiency over sequential or purely autoregressive approaches~\cite{sun2024emu2}.

\textbf{Efficiency Analysis.} Our framework reduces the DiT's computational complexity from quadratic to linear growth with the number of images~\cite{huang2024iclora, huang2024gdt,cai2025moc,dalal2025ttt,xiao2025captain}, an advantage we quantitatively analyze in the appendix (\cref{sec:efficiency_analysis}). This is achieved by using an MLLM mediator and learnable queries to encode inter-image coherence. This design also shifts the bottleneck to the highly-optimizable MLLM component~\cite{dao2022flashattention, pytorch_flexattention, unsloth_ai}, further improving efficiency. At inference, this enables parallel planning and generation, surpassing sequential approaches~\cite{sun2024emu2}.

\subsection{Progressive Training}
\label{sec:training}

We design a multi-stage progressive training strategy, enabling Narrative Weaver to gradually master narrative planning (Stage 1), semantically coherent visual generation (Stage 2), and fine-grained consistent visual generation (Stage 3). This approach is particularly effective under computational and data constraints. Upon completion, the trained model seamlessly integrates language modeling, visual understanding, and viusal generation within a unified multimodal framework (\cref{fig:method} (a)).

\textbf{Stage 1: Narrative Planning.} In this initial stage, we train the MLLM component while keeping the ViT encoder frozen. The model learns to formulate narrative plans and determine optimal timings for visual generation. Notably, our carefully designed attention mask significantly accelerates this training phase. The training objective is to minimize the negative log-likelihood of the ground-truth narrative text tokens, employing a standard cross-entropy loss for next-token prediction:
\begin{equation}
\mathcal{L}_{\text{narrative}} = - \sum_{j=0}^{N_T-1} \log P(t_{j,\text{gt}} | \bm{I}, \{t_{k,\text{gt}}\}_{k=0}^{j-1}),
\end{equation}
where $t_{j,\text{gt}}$ represents the $j$-th ground-truth token in a narrative sequence of length $N_T$, conditioned on the input $\bm{I}$ and all preceding ground-truth text tokens.

\textbf{Stage 2: Semantically Coherent Visual Generation.}
This stage focuses on training the learnable queries and the projector connecting the MLLM to the diffusion model, aiming to align the queries with the diffusion model's semantic space. We first pre-train on 30M large-scale, publicly available low-resolution (256$\times$256) text-image pairs. This is followed by fine-tuning on 60K high-resolution (512$\times$512) curated samples~\cite{chen2025blip3}. Subsequently, interleaved text-visual sequences are employed to facilitate learning attention patterns over relevant historical text and visual information. Training then proceeds with a standard Flow Matching objective, which guides the diffusion model to effectively generate visual content conditioned on our derived signals:
\begin{equation}
\mathcal{L}_{\text{visual}} = \mathbb{E}_{t, x_0, \epsilon, q_n} \left[ \| \mathbf{v}_{\theta}(x_t, t, q_n) - (\epsilon - x_0) \|^2 \right].
\end{equation}
Here, $x_t = (1-t)x_0 + t\epsilon$ is the noisy latent, $x_0$ is the ground-truth VAE feature, $\epsilon$ is Gaussian noise, $\mathbf{v}_{\theta}$ is the predicted vector field, and $q_n$ is the learnable query for n-th visual output.

% \textbf{Stage 2: Instruction Comprehension.} This stage leverages large-scale publicly available text-image pair datasets to train only the Learnable Query parameters and the projector connecting MLLM to the diffusion model. The objective is to align the Learnable Query with the semantic space of the diffusion model. To optimize training efficiency, we first pre-train on 30M text-image pairs at low resolution (256$\times$256), followed by high-resolution (512$\times$512) fine-tuning on 60K curated high-quality samples~\cite{chen2025blip3}.

% \textbf{Stage 3: Semantic Coherence.} While still updating only the Learnable Query and projector, this stage employs interleaved text-keyframe sequences. This enables the model to learn attention patterns over relevant historical text and visual information. After this phase, the model achieves preliminary semantic-level coherence while still lacking fine-grained consistency.

\textbf{Stage 3: Fine-grained Alignment.} 
In this final stage, we fully train the diffusion model aiming for fine-grained inter-visual consistency. To achieve this, the diffusion model's conditioning signal is augmented to integrate low-level conditional visual features $f^{\text{cond}}$ (derived from a VAE branch) and features from preceding visual outputs $\{\hat{f}_i\}_{i=1,2,\dots}$ (supplied by the Memory Bank). These additional features are then combined with the learnable query $q_n$, forming the comprehensive conditioning signal $\mathbf{C}_n$. The training objective remains the Flow Matching loss:
\begin{equation}
\mathcal{L}_{\text{visual}} = \mathbb{E}_{t, x_0, \epsilon, \mathbf{C}_n} \left[ \| \mathbf{v}_{\theta}(x_t, t, \mathbf{C}_n) - (\epsilon - x_0) \|^2 \right],
\end{equation}
where $\mathbf{C}_n$ is comprehensive conditioning signal (\cref{sec:model}).

% This multi-stage, decoupled training scheme decomposes the target capabilities into distinct stages, allowing the model to progressively advance from basic to advanced competencies. 
% \textbf{Stage 1} and \textbf{Stage 2} significantly reduce data dependence on structured \texttt{[text]-[image]-[text]-[image]} sequences, a data format scarce in public resources, while still leveraging large-scale, easily accessible datasets to incrementally enhance essential model abilities.

\begin{figure*}[!ht]
  \centering
    \includegraphics[width=.98\linewidth]{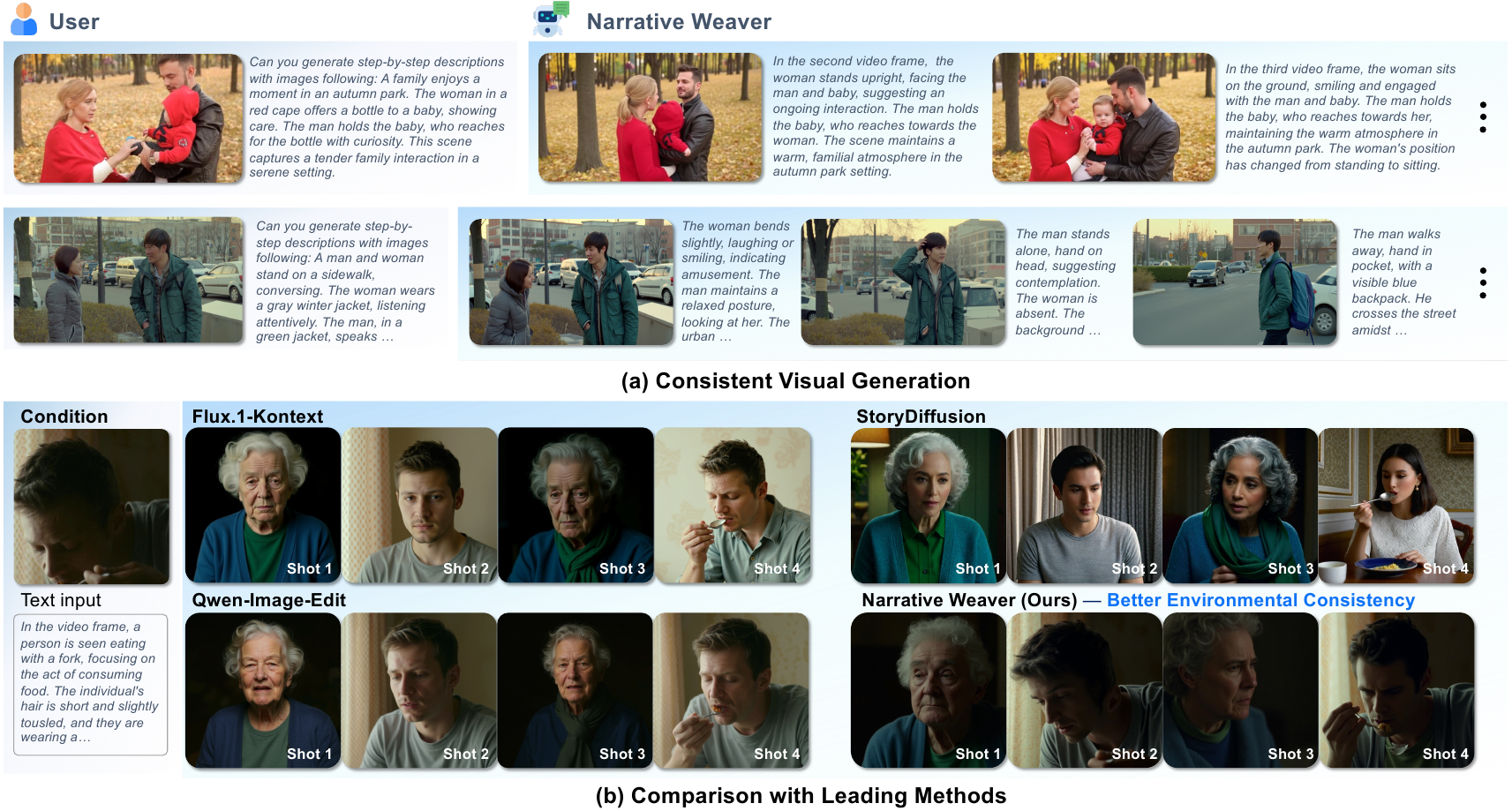}
    \vspace{-3mm}
    \caption{\textbf{Qualitative results of consistent visual generation.} (a) Narrative Weaver produces visually coherent frames that preserve both stylistic and semantic alignment with the given prompts, while effectively advancing the cinematic story progression. (b) Our model maintains environmental consistency conditioned on the input image and achieves more natural visual transitions compared to other methods.}
  % \caption{Visualization results of Narrative Weaver on key-frame generation for narrative consistency. The model produces visually coherent frames that maintain both style and semantic alignment with the prompt, while effectively driving forward the cinematic story progression. In addition, the generated transitions leverage classical filmmaking techniques (Shot/Reverse Shot, Cross Cutting) to enhance narrative clarity and continuity.}
  \label{fig:task1_1}
  \vspace{-5mm}
\end{figure*}

\section{Data Construction}
\label{sec:data}
% Recognizing keyframe-based methods as a viable pathway toward efficient long-video generation, we observe a notable lack of high-quality public datasets in this domain. To bridge this gap, we present a comprehensively annotated long-video keyframe dataset in the e-commerce domain, demanding strict subject consistency with reference images. The dataset features coherent narrative scripts, well-structured formatting, high-resolution and aesthetically selected keyframes, and strong inter-frame consistency. In this section, we detail the pipeline for \textbf{prompt generation}, \textbf{image generation}, and \textbf{data filtering}.

% Effective automatic planning and generation of long-range visual content requires datasets that support conditional image-to-multiframe generation with coherent narrative planning. However, existing datasets exhibit significant limitations: large-scale pretraining corpora from OmniGen2~\cite{wu2025omnigen2} suffer from inconsistent quality and lack conditional image inputs; annotated video snippet collections such as CI-VID~\cite{ju2025ci-vid} lack conditional image guidance and typically contain fewer than four clips per sample; while narrative-focused datasets like StoryStream~\cite{yang2025seed-story} remain confined to simplified animation scenarios. These limitations collectively underscore the pressing need for specialized datasets that bridge conditional visual grounding, narrative planning, and multi-frame consistency.

Effective automatic planning of long-range visual content requires datasets for conditional image-to-multiframe generation with coherent narrative structures. We introduce the E-commerce Advertising Video Storyboard Dataset (EAVSD) to meet this need, comprising $\sim$330K high-quality images. Each sample in EAVSD is a complete narrative instance, consisting of an initial condition (a product image and textual instruction) paired with its corresponding target output (a narrative plan and the resulting storyboard images). This structure provides multi-modal conditions, supports text-driven narrative planning, and ensures high inter-frame consistency. 
% Visual examples from EAVSD are provided in \cref{sec:Prompts}. 
Visual examples from EAVSD are provided in the supplementary material (\cref{sec:Prompts}).
In the remainder of this section, we detail the construction pipeline.

\textbf{Prompt Generation.} We first curate key selling points, product descriptions, and promotional texts from proprietary e-commerce sources. Using a locally deployed Qwen3-30B-A3B model~\cite{yang2025qwen3}, we then generate multiple detailed textual prompts for image synthesis. Each prompt contains comprehensive keyframe descriptions—covering product presentation, character interactions, scene composition, emotional tone, lighting conditions, color palette, and props—along with professional shot guidance including shot types, camera angles, and potential camera movements. Detailed prompt examples are provided in the supplementary material (\cref{sec:Prompts}).

\textbf{Image Generation.} This phase comprises reference image generation and subsequent keyframe synthesis. While several commercial models (\eg, SeedDream4.0~\cite{seedream2025seedream}) support multi-image generation, our evaluation reveals limitations in maintaining cross-frame consistency. To address this, we adopt a sequential generation pipeline: we first generate reference images using prompts from the previous stage, then leveraging the reference image alongside original product information to refine keyframe descriptions via LLM reasoning. Subsequent keyframes are synthesized through specialized image editing models. For this stage, Qwen-Image~\cite{wu2025qwenimage} and Flux.1-kontext~\cite{batifol2025kontext} are employed for reference image generation and subsequent frame synthesis, respectively. Extensive prompt engineering was applied throughout this process to ensure high-quality visual outputs, with detailed templates provided in the supplementary material (\cref{sec:Prompts}).

\textbf{Data Filter.} The images produced through the Image Generation pipeline undergo systematic quality filtering to ensure high standards. Specifically, reference images generated by Qwen-Image are filtered out if they contain AI artifacts such as malformed fingers, extra limbs, or other structural anomalies. Additionally, frames synthesized via Kontext are evaluated for consistency with reference images in both entity preservation and stylistic coherence. The entire filtering process is automated using our locally deployed Qwen2.5-VL-32B~\cite{bai2025qwen25vl}. 
% 目前已知问题时会出现AI幻觉（如手指不完整，多条腿/手臂等）还有需要需要单独筛选图片中出现镜子里人物（有镜子时约等于多人图片，会导致kontext识别错误模特

\section{Experiments}
\label{sec:experiments}

In this section, we systematically evaluate Narrative Weaver by addressing three pivotal research questions that target its core capabilities in long-range visual generation:

\begin{itemize} 
\item \textbf{Q1:} To what degree can Narrative Weaver maintain consistency when generating long-form visual content? 
\item \textbf{Q2:} How well does it translate a high-level narrative plan into a coherent visual sequence? 
\item \textbf{Q3:} Is Narrative Weaver effective and practical for real-world content creation tasks? 
\end{itemize}

\begin{table}[!t]
    \caption{GPT-4o Evaluation of Consistent Visual Generation.}
    \label{tab:task1_gpt-4o}
    \vspace{-3mm}
    \centering
    \resizebox{0.47\textwidth}{!}{
    \begin{tabular}{l|cc|ccccc}
        \toprule
         \multirow{2}{*}{\textbf{Method}}&  \multicolumn{2}{c|}{\textbf{Capability}} & \multicolumn{5}{c}{\textbf{GPT-4o Score} $ \uparrow $ }\\
          & Text. & Ctrl.  & ITC & RGC & MSSC & MSCC & IMQ  \\
         \midrule
         TALC~\cite{bansal2024talc}  & \ding{55} & \ding{55}  & 2.87 & 1.86 & 6.94 & 5.81 & 3.20 \\ 
         StoryDiffusion~\cite{zhou2024storydiffusion}  & \ding{55} & \ding{55}  & 6.54 & 5.86 & 7.48 & 6.00 & 6.80   \\
         IP-Adapter~\cite{ye2023ip-adapter}  & \ding{55} & \ding{51}  & 7.11 & 6.10 & 8.57 & 7.57 & 6.65\\
         AnimeShooter~\cite{qiu2025animeshooter}  & \ding{55} & \ding{51}  & 2.80 & 2.39 & 4.98 & 4.19 & 4.24 \\
         Flux.1-kontext~\cite{batifol2025kontext}  & \ding{55} & \ding{51}  & 7.06 & \textbf{9.41} & 8.11 & 7.28 & 6.94\\
         Qwen-Image-Edit~\cite{wu2025qwenimage}  & \ding{55} & \ding{51}  & 7.46 & 7.44 & 8.43 & 7.81 & 7.29  \\
         \rowcolor{blue!10}
         \textbf{Narrative Weaver}  & \ding{51} & \ding{51}  & \textbf{7.54} & 8.86 & \textbf{8.67} & \textbf{7.91} & \textbf{7.35}  \\
        \bottomrule
    \end{tabular}
    }
    \vspace{-5mm}
\end{table}

\subsection{Experimental Setup}

% It should be noticed that Narrative Weaver is a general-purpose visual-generation framework that can be applied directly to long-range, consistency-sensitive image or video generation. However, due to computational constraints and evaluation feasibility, we perform experiments and quantitative evaluation in the keyframe image generation setting for long videos.
Narrative Weaver is a general framework applicable to both long-range image and video generation. To enable rigorous and feasible quantitative evaluation of long-range consistency, we benchmark our method on the task of generating coherent keyframes for long-form visual narratives.

Our implementation is built upon Qwen2.5-VL-3B~\cite{bai2025qwen25vl} as the MLLM backbone for narrative planning and Flux.1-Dev~\cite{flux1dev2024} for visual generation. 
Our multi-stage training strategy is applied to each dataset as follows: Stages 1 and 2 are trained for 3 epochs to establish text planning and coarse-grained alignment capabilities. Stage 3 is then trained for an additional 1-2 epochs to refine fine-grained consistency.
% We apply a multi-stage training strategy: Stages 1 and 2 run for 3 epochs to establish coarse-grained alignment, while Stage 3 runs for an additional 1-2 epochs to refine fine-grained consistency. 
For complete reproducibility, all hyperparameters and further implementation details are provided in supplementary material \cref{sec:experimental_details}. All training is conducted on 8 GPUs (global batch size 8). We leverage PyTorch's FlexAttention~\cite{pytorch_flexattention} to accelerate attention computation by nearly 2$\times$, optimizing training efficiency.

\subsection{Evaluating the Narrative Weaver}
% Q1 解释图文对齐First, achieving precise control is difficult, as per-shot instructions can be “diluted” within the context of the entire prompt. 主要评估图像与参考图像之间的一致性，一旦扩展到整个影片序列的一致性（LLM评估）我们的模型优势就被放大

\subsubsection{Consistent Visual Generation (Q1)}

\textbf{Evaluation Protocol.} We curated a test set of approximately 627k multi-keyframe samples derived from diverse video sources in the OmniGen2 dataset~\cite{wu2025omnigen2}. Performance was assessed using a combination of automated metrics (CLIP Score~\cite{radford2021clip}, DreamSim~\cite{fu2023dreamsim}) and an LLM-based evaluation (GPT-4o~\cite{openai_gpt4o_2024}). These metrics were chosen to holistically measure cross-frame coherence, reference-frame alignment, and text-image matching (details in supplementary material \cref{sec:eva_details}).

\textbf{Baselines.} We benchmark Narrative Weaver against a comprehensive suite of open-source methods, categorized by their approach: chunk-based methods (TALC~\cite{bansal2024talc}, AnimeShooter~\cite{qiu2025animeshooter}), keyframe-based strategies (StoryDiffusion~\cite{zhou2024storydiffusion}, IP-Adapter~\cite{ye2023ip-adapter}), and leading image editing models (Flux.1-kontext~\cite{batifol2025kontext}, Qwen-Image-Edit~\cite{wu2025qwenimage}).

\textbf{LLM-based Quantitative Evaluation.} The results of our LLM-based evaluation are presented in \cref{tab:task1_gpt-4o}. This assessment covers five key dimensions: Image-Text Consistency (ITC), Reference-Generated Consistency (RGC), Multi-Shot Style Consistency (MSSC), Multi-Shot Content Consistency (MSCC), and Image Quality (IMQ). To enhance evaluation reliability, we implemented an ``Analyze-then-Judge" Chain-of-Thought (CoT) prompting strategy~\cite{wei2022cot}, yielding assessments more aligned with human judgment. Narrative Weaver achieves state-of-the-art performance across all dimensions except RGC, where it is surpassed only by specialized editing models that are explicitly optimized for reference fidelity. Detailed prompts and implementation specifics for baseline comparisons are provided in supplementary material \cref{sec:eva_details} and \cref{sec:experimental_details}.

\begin{figure}[!t]
  \centering
    \includegraphics[width=.95\linewidth]{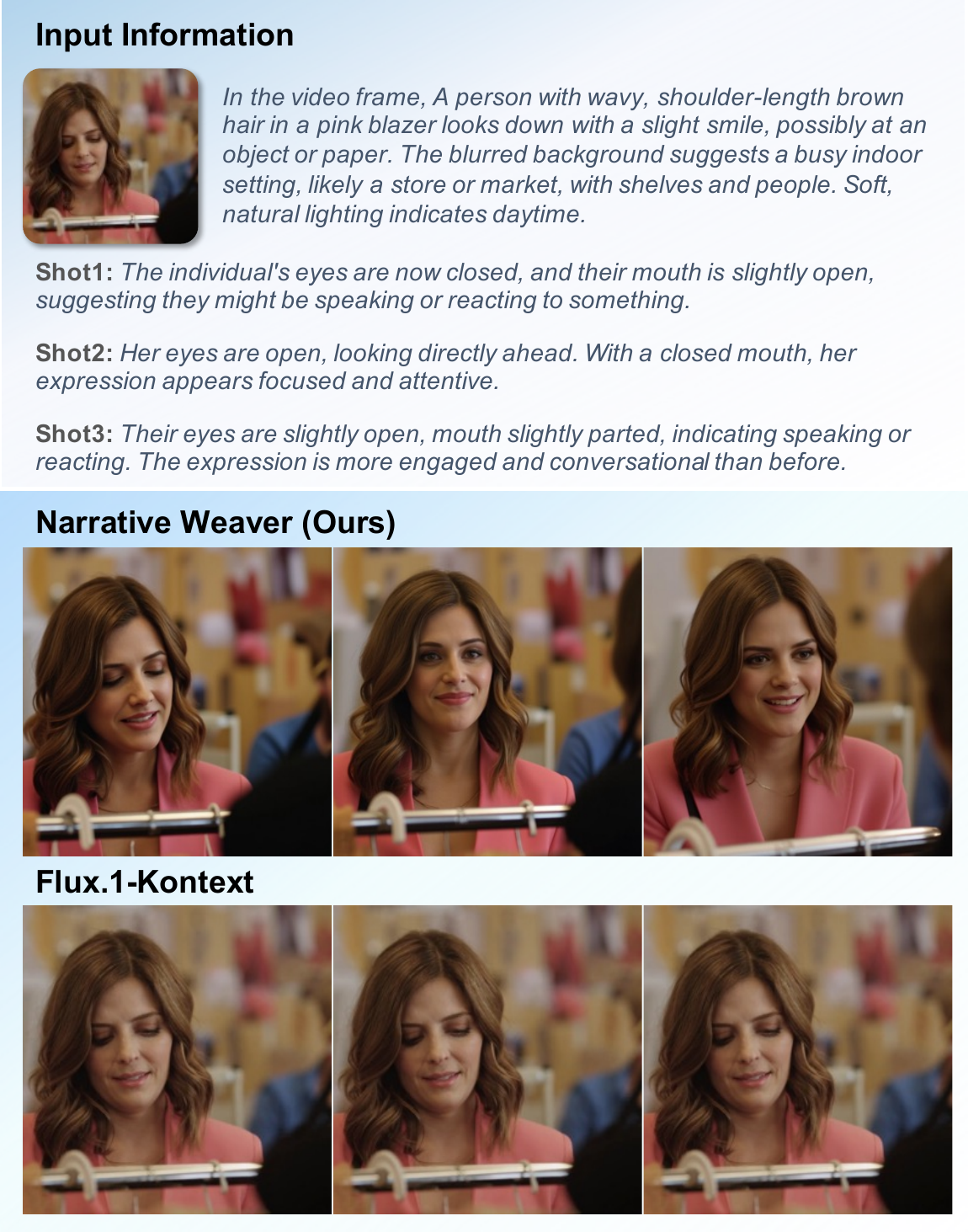}
    \vspace{-3mm}
  \caption{Flux.1-Kontext tend to exhibit \colorbox{red!20}{“copy–paste”} behavior when failing to interpret instructions, resulting in a misleading appearance of high consistency.}
  \label{fig:task2_copy-paste}
  \vspace{-3mm}
\end{figure}

\begin{figure}[!t]
  \centering
    \includegraphics[width=.95\linewidth]{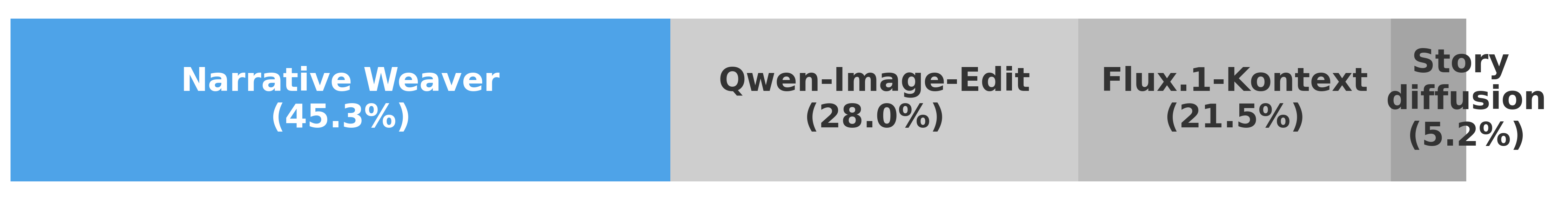}
    \vspace{-3mm}
  \caption{User Study Results: Model Preference Distribution. The results were aggregated from over 180 responses, each representing user's selection of the most preferred output. }
  \label{fig:userstudy}
  \vspace{-3mm}
\end{figure}

\textbf{Automated Evaluation.} Automated metrics (\cref{tab:task1_clip-dreamsim}) corroborate our findings. Narrative Weaver outperforms all multi-scene video generation baselines on CLIP Score and DreamSim. While surpassed by specialized editing models, this is an expected outcome, as these metrics reward frame similarity over inter-frame consistency. In contrast, Flux.1-Kontext often shows ``\textit{copy-paste}" artifacts or static behavior (\cref{fig:task2_copy-paste}). Our method avoids this by balancing consistency with dynamic storytelling. To validate this qualitative advantage, which is overlooked by automated metrics, we conducted a user study. The results (\cref{fig:userstudy}) confirm a strong user preference for our model.

% \textbf{Automated Evaluation.} The automated metrics (\cref{tab:task1_clip-dreamsim}) corroborate our findings. Narrative Weaver outperforms all multi-scene video generation baselines on CLIP Score and DreamSim. While surpassed by specialized editing models, this is an expected outcome, as these metrics reward frame similarity over inter-frame consistency. In contrast, Flux.1-Kontext often shows \underline{copy-paste} artifacts or static behavior (\cref{fig:task2_copy-paste}). Our method avoids this by successfully balancing consistency with dynamic storytelling, a qualitative advantage confirmed by an additional user study (\cref{fig:userstudy}).

% \textbf{Automated Evaluation.} The automated metrics, shown in \cref{tab:task1_clip-dreamsim}, corroborate our findings. Narrative Weaver outperforms all multi-scene video generation baselines on both CLIP Score and DreamSim. While it is surpassed by specialized editing models, this is an expected outcome, as these metrics primarily reward strict reference-frame similarity rather than nuanced inter-frame consistency and narrative progression. Notably, Flux.1-Kontext often exhibit \underline{copy-paste} artifacts or static behavior on complex prompts, whereas our method successfully balances consistency with dynamic storytelling. (\cref{fig:task2_copy-paste})
% \colorbox{red!20}{copy-paste}
\begin{table}[!t]
    \caption{Automated Evaluation (CLIP Score / DreamSim) of Consistent Keyframe Generation (Q1).}
    \label{tab:task1_clip-dreamsim}
    \vspace{-3mm}
    \centering
    \resizebox{0.45\textwidth}{!}{
    \begin{tabular}{lcccc c}
    \toprule
         \multirow{2}{*}{\textbf{Method}}& \multicolumn{4}{c}{Shot-level} & Story-level\\
         \cmidrule(lr){2-5}\cmidrule(lr){6-6}
         & shot-1 & shot-2 & shot-3 & shot-4 & Avg. \\
        \midrule
         \multicolumn{6}{c}{\textbf{DreamSim} $ \downarrow $ } \\
         \Xhline{0.2pt} % 细线
         TALC~\cite{bansal2024talc}  & 83.60 & 84.03 & 88.32 & 89.38 & 83.87\\ 
         StoryDiffusion~\cite{zhou2024storydiffusion}  & 54.63 & 57.35& 59.06& 58.90& 56.33\\
         IP-Adapter~\cite{ye2023ip-adapter}  & 33.03 & 33.57 & 34.07& 34.83& 33.30\\
         AnimeShooter~\cite{qiu2025animeshooter}  & 75.89 & 71.49 & 71.87& 70.29& 73.14\\
         \rowcolor{gray!30} 
         Flux.1-kontext~\cite{batifol2025kontext}  & 4.75 & 3.08 & 3.46& 3.22& 3.71\\
         Qwen-Image-Edit~\cite{wu2025qwenimage}  & \underline{14.19} &  \underline{13.39}& \underline{13.03}& \underline{12.10}& \underline{13.78}\\
         \textbf{Narrative Weaver (Ours)}  & \textbf{12.69} & \textbf{11.75} & \textbf{11.45} & \textbf{10.83}& \textbf{12.18}\\
         \hline
         \multicolumn{6}{c}{\textbf{CLIP Score} $ \uparrow $ } \\
         \Xhline{0.2pt} % 细线
         TALC~\cite{bansal2024talc}  & 50.60 & 50.71 & 47.23 & 47.77&50.54 \\ 
         StoryDiffusion~\cite{zhou2024storydiffusion}  & 64.25& 60.68 & 59.60 & 58.09 & 62.20 \\
         IP-Adapter~\cite{ye2023ip-adapter}  & 83.18& 81.88& 80.70& 80.77 & 82.32 \\
         AnimeShooter~\cite{qiu2025animeshooter}  & 54.81& 55.82& 54.94 & 56.88 & 55.80\\
         \rowcolor{gray!30} 
         Flux.1-kontext~\cite{batifol2025kontext}  & 96.43 & 97.50& 97.32 & 97.78 & 97.17 \\
         Qwen-Image-Edit~\cite{wu2025qwenimage}  & \textbf{90.30} & \textbf{91.24} & \textbf{91.07} & \textbf{92.15} & \textbf{91.19} \\
         \textbf{Narrative Weaver (Ours)}  & \underline{89.32} & \underline{90.54} & \underline{90.75} & \underline{91.70} & \underline{89.98} \\
        \bottomrule
    \end{tabular}
    }
    \vspace{-6mm}
\end{table}

\textbf{Qualitative Analysis.} Qualitative results in \cref{fig:task1_1} (a) provide visual evidence of Narrative Weaver's superior performance. Our framework not only maintains robust style consistency and character identity across frames but also executes precise temporal progressions that align with the narrative instructions. 
A direct comparison against three leading methods in \cref{fig:task1_1} (b) further highlights its advantages: Narrative Weaver uniquely maintains environmental consistency, while all baselines falter in preserving lighting conditions. Notably, the consistency between non-adjacent frames such as Shot 1 and Shot 3 confirms that our learnable query effectively enables long-range information exchange within the model.

\begin{figure*}[!ht]
  \centering
    \includegraphics[width=.98\linewidth]{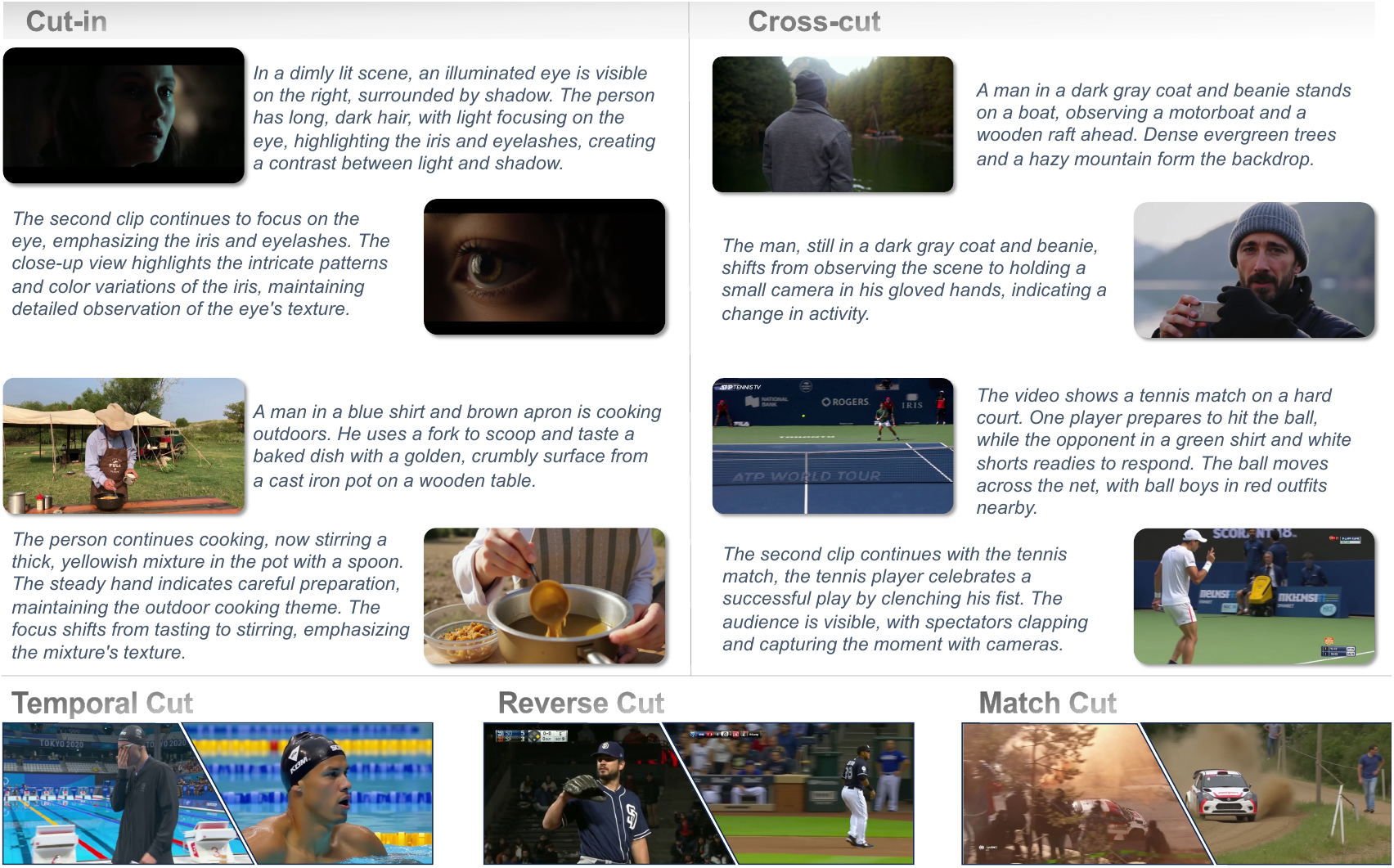}
    \vspace{-2mm}
    \caption{\textbf{Qualitative results of autonomous narrative planning.} Narrative Weaver demonstrates a dual capability: maintaining robust visual consistency while also employing fundamental cinematic techniques. The figure showcases examples where the model autonomously plans and generates contextually appropriate subsequent shots that adhere to standard conventions, including cut-ins for detail, cross-cuts for parallel action, and so on.}
  % \caption{Visualization results of Narrative Weaver on key-frame generation for Narrative Planning. Our model demonstrates both consistent keyframe generation and emergent understanding of cinematic techniques, autonomously planning and visualizing appropriate subsequent shots.}
  \label{fig:task2_1}
  \vspace{-5mm}
\end{figure*}

% \begin{table*}[!ht]
%     \caption{GPT-4o evaluation on the autonomous narrative planning capability of Narrative Weaver (Q2).}
%     \label{tab:task2}
%     \vspace{-2mm}
%     \centering
%     \resizebox{0.95\textwidth}{!}{
%     \begin{tabular}{l|c|cccccc|cccccc}
%         \toprule
%            \multirow{2}{*}{\textbf{Method}} & \multirow{2}{*}{\textbf{Params.}} &  \multicolumn{6}{c|}{\textbf{Continuation Generation}} &  \multicolumn{6}{c}{\textbf{Question-based Generation}} \\
%           & & Style & Entity & Trend & CPL. & ImgQ & IRS & Style & Entity & Trend & CPL. & ImgQ & IRS \\
%          \midrule
%          MiniGPT-5~\cite{zheng2023minigpt} & 7B & 6.11 & 5.83 & 5.93 & 6.53 & 5.80 & 2.79 & 6.62 & 5.96 & 5.83 & 6.26 & 5.67 & 2.73\\ 
%          SEED-Llama 8B~\cite{ge2023seed-llama} & 8B & 6.64 & 6.57 & 6.28 & \textbf{6.71} & 6.07 & \textbf{3.14} & 7.35 & 5.00 & 3.85 & 5.07 & 5.21 & 2.57\\
%          SEED-Llama 14B~\cite{ge2023seed-llama}  & 14B & 6.71 & 5.71 & 5.50 & 6.14 & 5.85 & 2.57 & 7.00 & 6.28 & 5.71 & 6.02 & 5.85 & \textbf{3.28}\\
%          Qwen2.5-VL + Flux.1-Dev  & 15B & 6.45 & 5.86 & 5.66 & 5.81& 5.48 &2.94 &5.39 & 4.89 & 4.69 & 4.79 & 4.73 & 2.98\\
%          \rowcolor{blue!10}
%          \textbf{Narrative Weaver (Ours)} & 15B & \textbf{7.07} & \textbf{7.21} & \textbf{6.77} & 6.00 & \textbf{7.29} & 2.64 &  \textbf{7.10} & \textbf{6.49} & \textbf{6.67} & \textbf{7.14} & \textbf{6.54} & 2.77\\
%         \bottomrule
%     \end{tabular}
%     }
% \end{table*}

\subsubsection{Autonomous Narrative Planning (Q2)}
\textbf{Evaluation Protocol.} For Q2, we evaluated Narrative Weaver's autonomous narrative planning capabilities. We adopted the narrative-intensive ``Question-based Generation" task from the CoMM benchmark~\cite{chen2025comm} as the primary testbed. This task requires the model to generate alternating text-image sequences to respond to queries. 
To ensure the rigor and reproducibility of our evaluation, we first curated a validated subset of the CoMM test set, addressing minor inconsistencies, such as invalid data URLs. All baselines were then re-evaluated on this standardized subset using the officially provided checkpoints, guaranteeing a fair comparison.
Furthermore, we established a baseline by combining Qwen-2.5VL-3B with FLUX.1-Dev, serving as a direct counterpart to Narrative Weaver. To further assess planning within cinematic contexts, we also employed the CI-VID dataset~\cite{ju2025ci-vid} to examine the model's ability to generate coherent subsequent shots.

% \begin{table}[!ht]
%     \caption{GPT-4o evaluation on the autonomous narrative planning capability of Narrative Weaver (Q2).}
%     \label{tab:task2}
%     \vspace{-2mm}
%     \centering
%     \resizebox{0.47\textwidth}{!}{
%     \begin{tabular}{l|c|cccccc}
%         \toprule
%            \multirow{2}{*}{\textbf{Method}} & \multirow{2}{*}{\textbf{Params.}} &  \multicolumn{6}{c}{\textbf{Question-based Generation}} \\
%           & & Style & Entity & Trend & CPL. & ImgQ & IRS \\
%          \midrule
%          MiniGPT-5~\cite{zheng2023minigpt} & 7B &  6.62 & 5.96 & 5.83 & 6.26 & 5.67 & 2.73\\ 
%          SEED-Llama~\cite{ge2023seed-llama} & 8B &  7.35 & 5.00 & 3.85 & 5.07 & 5.21 & 2.57\\
%          SEED-Llama~\cite{ge2023seed-llama}  & 14B &  7.00 & 6.28 & 5.71 & 6.02 & 5.85 & \textbf{3.28}\\
%          Qwen2.5-VL + Flux.1-Dev  & 15B &5.39 & 4.89 & 4.69 & 4.79 & 4.73 & 2.98\\
%          \rowcolor{blue!10}
%          \textbf{Narrative Weaver (Ours)} & 15B & \textbf{7.10} & \textbf{6.49} & \textbf{6.67} & \textbf{7.14} & \textbf{6.54} & 2.77\\
%         \bottomrule
%     \end{tabular}
%     }
%     \vspace{-4mm}
% \end{table}
\begin{table}[!ht]
    \caption{GPT-4o evaluation on the autonomous narrative planning capability of Narrative Weaver (Q2).}
    \label{tab:task2}
    \vspace{-3mm}
    \centering
    \resizebox{0.47\textwidth}{!}{
    \begin{tabular}{l|cccccc}
        \toprule
           \multirow{2}{*}{\textbf{Method}} &  \multicolumn{6}{c}{\textbf{Question-based Generation}} \\
          &  Style & Entity & Trend & CPL. & ImgQ & IRS \\
         \midrule
         MiniGPT-5~\cite{zheng2023minigpt} &  6.62 & 5.96 & 5.83 & 6.26 & 5.67 & 2.73\\ 
         SEED-Llama-8B~\cite{ge2023seed-llama}  &  7.35 & 5.00 & 3.85 & 5.07 & 5.21 & 2.57\\
         SEED-Llama-14B~\cite{ge2023seed-llama}   &  7.00 & 6.28 & 5.71 & 6.02 & 5.85 & \textbf{3.28}\\
         Qwen2.5-VL + Flux.1-Dev   &5.39 & 4.89 & 4.69 & 4.79 & 4.73 & 2.98\\
         \rowcolor{blue!10}
         \textbf{Narrative Weaver (Ours)}  & \textbf{7.10} & \textbf{6.49} & \textbf{6.67} & \textbf{7.14} & \textbf{6.54} & 2.77\\
        \bottomrule
    \end{tabular}
    }
    \vspace{-6mm}
\end{table}
\textbf{Quantitative Results.} As presented in \cref{tab:task2}, we conducted a quantitative comparison on the CoMM benchmark across multiple dimensions. The metrics include Style, Entity, and Trend to measure consistency; CPL (Completeness) to assess narrative integrity; ImgQ (Image Quality) to evaluate visual fidelity; and IRS (Illustration Relevance Score) to quantify text-image alignment. Narrative Weaver achieves the best overall performance, particularly excelling in multi-faceted consistency, narrative completeness, and image quality. These results demonstrate its superior capability in seamlessly coordinating high-level narrative planning with high-fidelity visual generation.

\begin{figure*}[!ht]
  \centering
    \includegraphics[width=.98\linewidth]{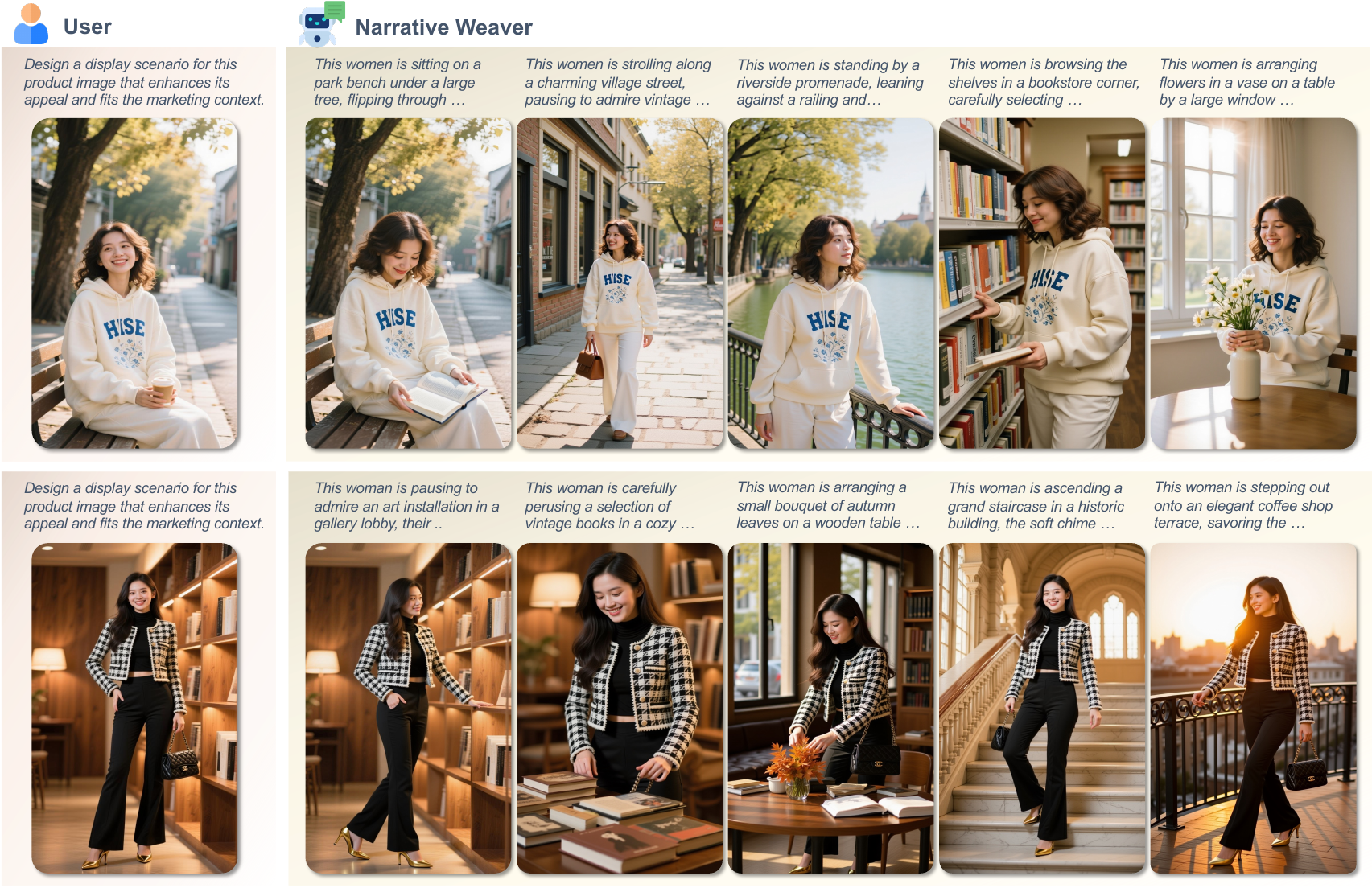}
    \vspace{-2mm}
    \caption{\textbf{Visual Scenery Planning for E-commerce Video Ads.} Visualization results demonstrate Narrative Weaver's capability in generating consistent keyframe sequences with precise scene composition for e-commerce scenarios.}
  \label{fig:task3}
  \vspace{-6mm}
\end{figure*}

\textbf{Qualitative Analysis of Narrative Planning.} Qualitative results, evidenced in \cref{fig:task2_1}, reveal that training in video keyframe data (CI-VID) equips our model with a practical understanding of cinematic language while maintaining generative consistency. The examples demonstrate Narrative Weaver's capacity to autonomously plan contextually appropriate subsequent shots, effectively employing standard cinematographic conventions. These include using cut-in shots for detail emphasis, reverse-shot sequences to maintain dialogue continuity, and cross-cutting to develop parallel narratives. 
This ability to orchestrate diverse shot types while preserving a coherent narrative flow validates the effectiveness of our methodology in integrating visual generation with fundamental storytelling principles.

\subsubsection{Extended  Application Scenarios (Q3)}
To evaluate practical utility, we tested Narrative Weaver in e-commerce advertising—a domain demanding strict visual consistency and narrative planning. On our EAVSD dataset, we tasked the model with generating storyboards from user instructions and product images. As shown in \cref{fig:task3}, 
% our framework generates coherent keyframe sequences, which form the basis for final videos available in our supplementary material. 
our framework generates coherent keyframe sequences that serve as the foundation for the final advertisement video.
This process involves designing contextually appropriate scenes while maintaining strict product identity. To validate this consistency, supplementary material provides a direct comparison with a leading model, confirming our method's superior performance.

\subsection{Ablation Study}
To validate the key components for Narrative Weaver's controllability and efficiency, we conducted an ablation study on our multi-stage training: Stage 2 (semantic coherence) and Stage 3 (visual consistency). As shown in \cref{tab:Abalation}, removing either stage significantly degrades performance across almost all metrics on our e-commerce benchmark. This confirms both stages are critical for achieving state-of-the-art performance. 
To visually illustrate the impact of these components, supplementary material \cref{fig:ablation} showcases the significantly enhanced fine-grained control of the fully trained model compared to its ablated counterparts.

\begin{table}[!t]
    % \vspace{-2mm}
    \caption{Ablation study validating the contributions of Stage 2 (semantic coherence) and Stage 3 (fine-grained control).}
    % \caption{Ablation study demonstrating the advantages of Narrative Weaver in efficient training (vs. w/o Training Stage 2) and fine-grained control (vs. w/o Training Stage 4).}
    \label{tab:Abalation}
    \vspace{-3mm}
    \centering
    \resizebox{0.47\textwidth}{!}{
    \begin{tabular}{cc|cccccc}
        \toprule
         stage2 & stage3 & ITC & RGC & MSSC & MSCC & IMQ & Avg.\\
         \midrule
         \ding{55} & \ding{55}  & 5.99 & 6.09 & 8.19 & 7.22 & 8.12 & 7.12  \\ 
         \ding{51} & \ding{55}  & 6.05 & 6.78 & 8.59 & 7.73 &  \textbf{8.20} & 7.47\\
         \ding{55} & \ding{51}  & 6.19 & 8.53 & 8.84 & 8.28 & 8.18  & 8.00\\ 
         \ding{51} & \ding{51}  &  \textbf{6.39} & \textbf{8.68} & \textbf{8.97} & \textbf{8.34} & 8.14 & \textbf{8.10} \\
        \bottomrule
    \end{tabular}
    }
    \vspace{-6mm}
\end{table}

\section{Conclusion}
\label{sec:conclusion}

% propose Narrative Weaver, 一个长视频生成框架集成了，细粒度控制、自主叙事能力以及相对高效的实现 with our 设计的多阶段训练策略。我们在实验部分分别对以上几点进行了验证。另外，意识到现有探索多场景长视频生成的公共数据稀缺并为了充分发掘方法的能力，我们在自己收集的电商领域数据下做了进一步验证实验，证实了方法的泛化性。 
% We present Narrative Weaver, an integrated long-range visual generation framework that achieves fine-grained controllability, autonomous narrative planning, and training efficiency through our carefully designed multi-stage training strategy. Extensive validation across diverse scenarios confirms the effectiveness of each key aspect of our approach. Furthermore, recognizing the scarcity of public datasets for exploring multi-scene long-video generation, we conduct additional experiments using our collected e-commerce dataset, demonstrating compelling generalization capability and practical applicability of the proposed framework in real-world settings. Due to computation limitation, only images generation are implemented in this work, however the proposed framework is universal, which can be applied to video clip generation directly, and this will be our future work.

We presented Narrative Weaver, a framework unifying fine-grained control and autonomous narrative planning with a data-efficient training strategy. Its effectiveness and real-world applicability were validated across diverse scenarios, including our new EAVSD benchmark. Though demonstrated on images, its architecture-agnostic design allows for a straightforward extension to video generation, which we leave as promising future work.

% We present Narrative Weaver, an integrated long-range visual generation framework that unifies fine-grained controllability, autonomous narrative planning, and training efficiency through a carefully designed multi-stage strategy. Comprehensive experiments across diverse scenarios validate the effectiveness of each component and demonstrate the robustness of our approach. To further address the lack of suitable benchmarks for long-horizon, multi-scene visual generation, we additionally evaluate our framework on a newly collected EAVSD dataset, showcasing strong generalization ability and promising real-world applicability.

% Due to current computational constraints, our experiments focus on image-based generation; nevertheless, the proposed framework is architecture-agnostic and can be directly extended to video clip generation, which holds even greater potential for practical deployment — a direction we leave for future work.

\section*{Acknowledgments}

This work was supported in part by the National Key Research and Development Program of China (2025YFA1805700); in part by the Capital's Funds for Health Improvement and Research (2026-1-2151); in part by the National Natural Science Foundation of China (82371112, 62501020); and in part by the Science Foundation of Peking University Cancer Hospital (JC202505).

{
    \small
    \bibliographystyle{ieeenat_fullname}
    \bibliography{main}
}
\clearpage
\setcounter{page}{1}
\maketitlesupplementary

This supplementary document provides additional details, visualizations, and analyses to support the claims and experiments presented in the main paper. Beyond expanding the empirical evidence, we also further clarify the methodological significance of Narrative Weaver.

Narrative Weaver is not merely an integration of existing components, but a systematic methodology designed to address the challenge of Long-Range Visual Consistency. Specifically, it bridges the gap between short-form visual generation and professional production workflows by enabling high-level narrative logic to consistently govern low-level visual details. Moreover, it establishes a transferable framework for long-range consistency: core components such as the \textit{Dynamic Memory Bank} and \textit{Dual-path Alignment} are modular and can be extended to related domains, including long-form video generation. Finally, we introduce a practical multi-stage progressive training strategy with customized attention mechanisms that efficiently decouple and align complex features, demonstrating strong empirical effectiveness even under resource constraints.

The remainder of this supplementary document is organized as follows:
% This supplementary document provides additional details, visualizations, and analyses to support the claims and experiments presented in the main paper. The material is organized as follows:

\begin{itemize}
    \item \textbf{Additional Details on Data Construction (\cref{sec:Prompts})} provides a deep dive into our novel data creation methodology. We detail the multi-step prompt engineering pipeline and describe the curation process for our new dataset.

    \item \textbf{Evaluation Details (\cref{sec:eva_details})} outlines the specifics of our evaluation framework. We detail the curation of our test sets, the precise implementation and setup for all baseline methods to ensure fair comparisons, and the methodology of our human evaluation study.

    \item \textbf{Experimental Details (\cref{sec:experimental_details})} is dedicated to the implementation and training of Narrative Weaver. We present the complete training recipe for our multi-stage strategy, including a detailed breakdown of all hyperparameters to ensure full reproducibility.

    \item \textbf{Additional Experimental Results (\cref{sec:Add_results})} presents an extensive gallery of additional qualitative results. This includes more visual examples from Narrative Weaver and numerous side-by-side comparisons against baselines to further substantiate our claims of superior consistency and aesthetic quality.

    \item \textbf{Detailed Ablation Results (\cref{sec:detail_ablation})} provides a quantitative analysis of the contribution of key components within our architecture. These detailed ablation studies validate our design choices and demonstrate the importance of each module.

    \item \textbf{Additional Efficiency Analysis (\cref{sec:efficiency_analysis})} presents a quantitative comparison of the computational cost of our proposed architecture against a vanilla self-attention baseline. 

    \item \textbf{Limitations and Future Works (\cref{sec:limitaions})} discusses the current limitations of our work and outlines promising directions for future research, including the extension to video generation and the need for broader dataset creation.
\end{itemize}

We believe these supplementary details will provide a comprehensive understanding of our work and its contributions.

\clearpage
\section{Additional Details on Data Construction}
\label{sec:Prompts}

\begin{figure}
  \centering
    \includegraphics[width=1.0\linewidth]{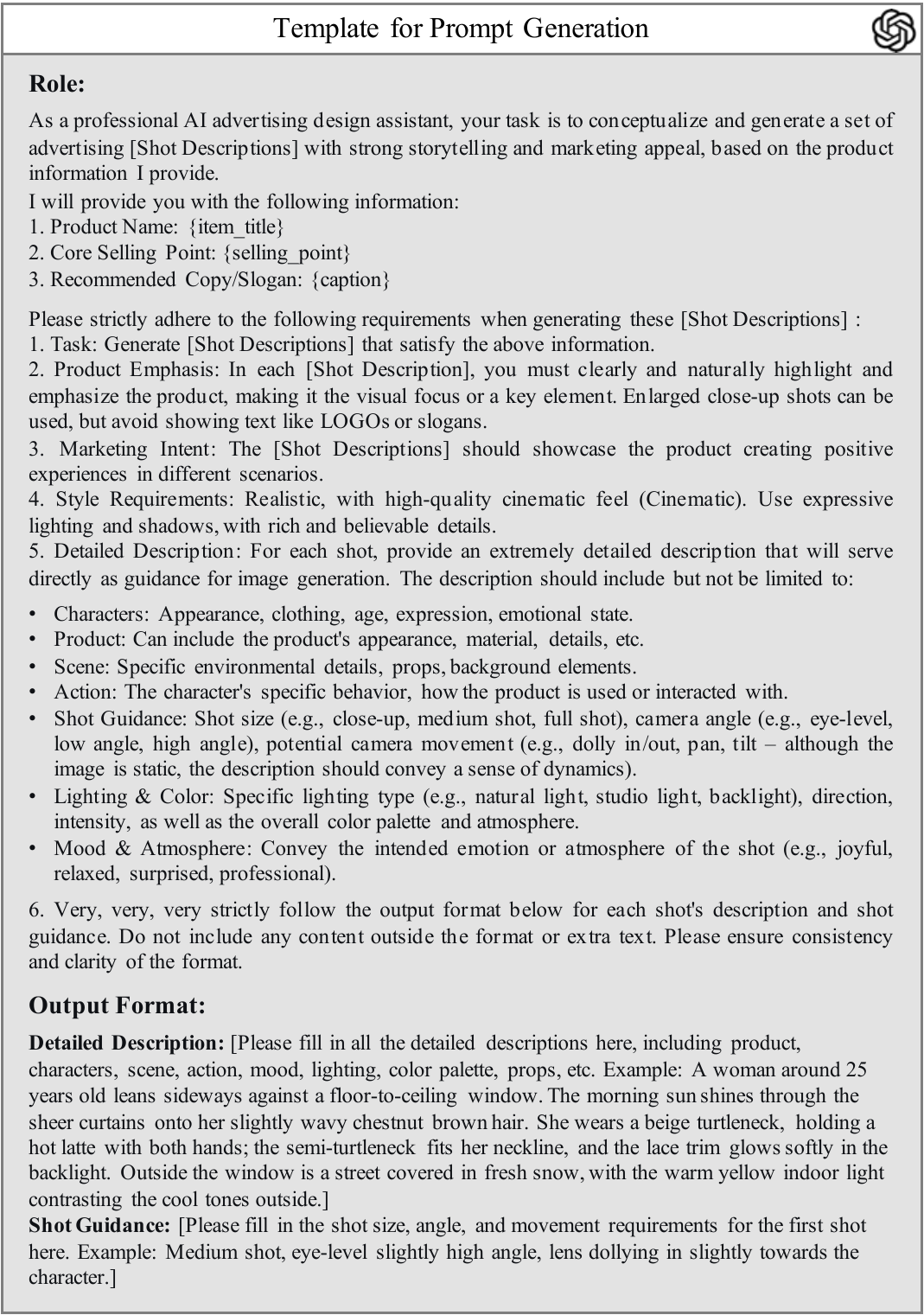}
  \caption{The prompt for Qwen3-30B-A3B to generate text instructions for reference image generation.}
  \label{fig:prompt1_supp}
\end{figure}

\begin{figure}
  \centering
    \includegraphics[width=1.0\linewidth]{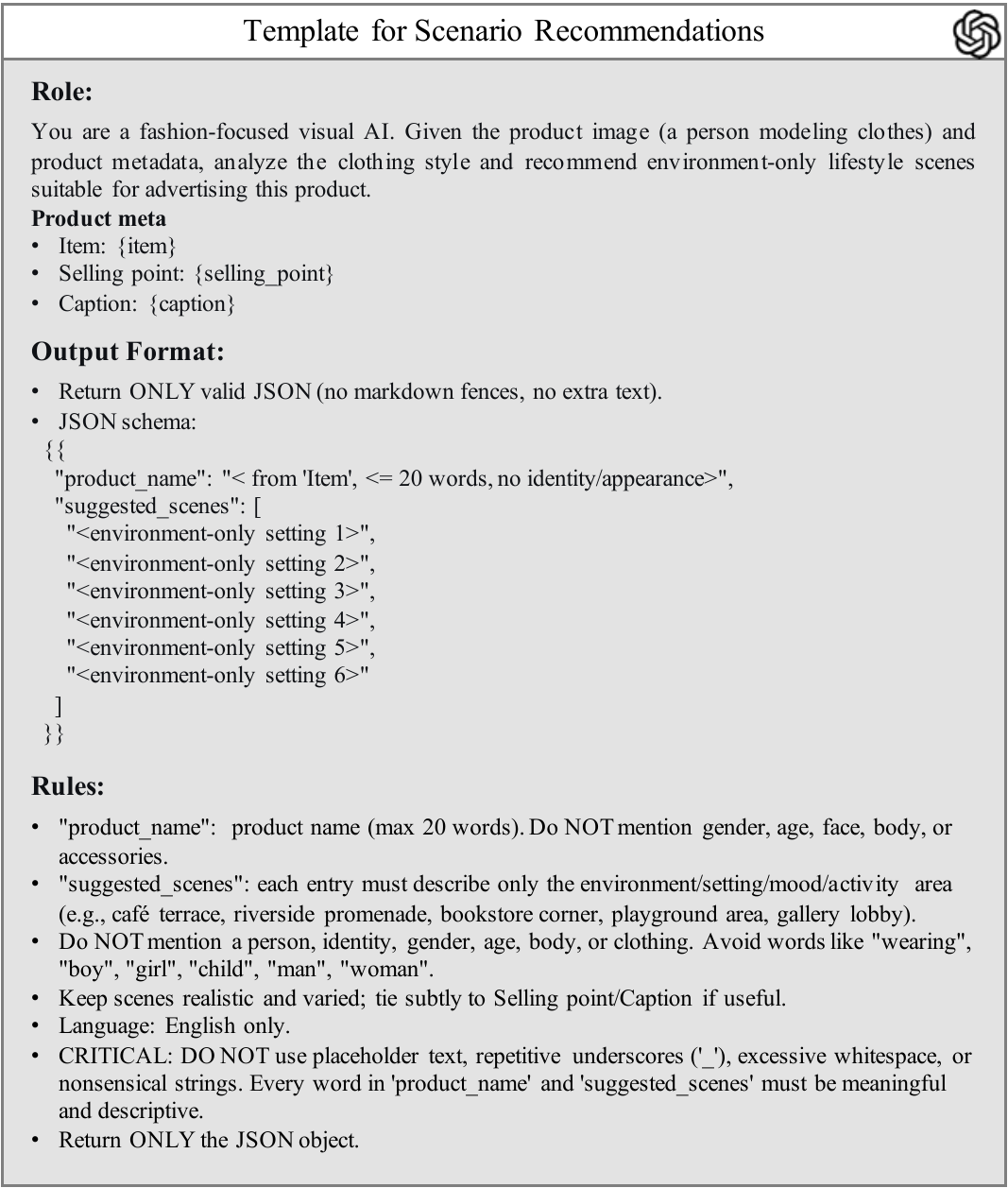}
  \caption{Prompt Template for Qwen3-30B-A3B in Generating Keyframe Scenario Recommendations.}
  \label{fig:prompt2_supp}
\end{figure}

\begin{figure}
  \centering
    \includegraphics[width=1.0\linewidth]{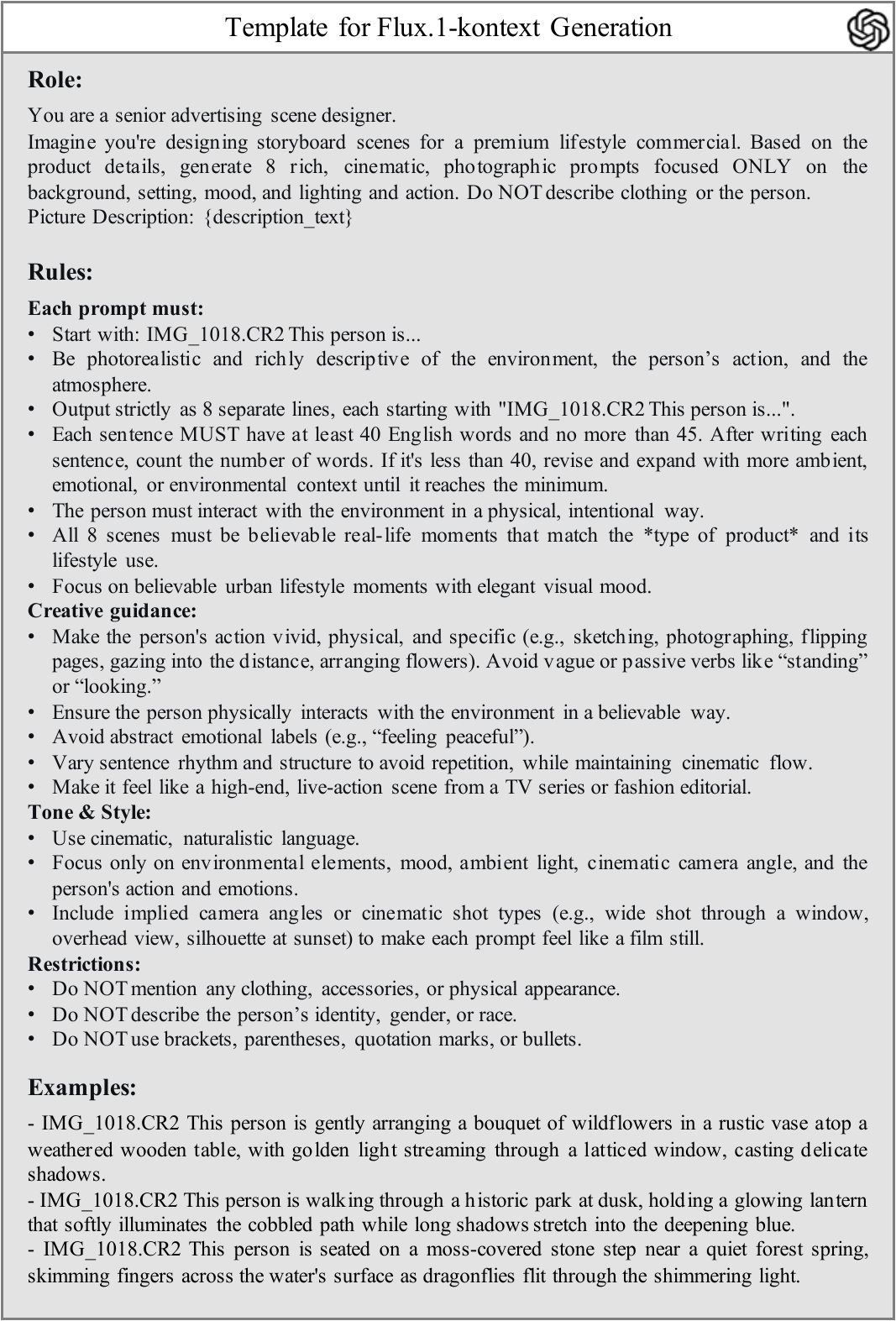}
  \caption{The prompt for Qwen3-30B-A3B to generate text instructions for Flux.1-kontext image generation.}
  \label{fig:prompt3_supp}
\end{figure}

% 在数据生成第二阶段——图像部分我们使用了LLM推理完成了：（1）场景推荐（可大幅提高后续生图场景真实性）以及（2）继续通过这个场景推荐的信息生成后续关键帧描述。我们发现把场景推荐单独提取出来作为子任务的策略是必要的，不贴合的场景会产生不真实的图片，类似于直接p图的效果
% 注意不要让模型描述人物特征，不描述人物和服装具体特征，着重于场景交互与背景描述。任何对于人物的描述都会导致后续编辑模型(Flux.1-kontext)生成与输入模特图产生偏差的图片
This section elaborates on the prompt generation methodology outlined in the main text. Our image generation pipeline involves three distinct LLM calls to sequentially produce: (1) shot descriptions for reference images, (2) scenario recommendations for subsequent keyframes, and (3) detailed captions for keyframe synthesis. We provide carefully engineered prompt templates in this section (\cref{fig:prompt1_supp}, \cref{fig:prompt2_supp}, \cref{fig:prompt3_supp}), which are being released to support community research. Our workflow is divided into two main stages:

\textbf{Stage 1: Reference Image Generation.} Our process begins with a dataset of 33,000 real-world product listings, encompassing product names, selling points, and recommendations across categories such as apparel, accessories, home goods, and bags. This information is fed into our self-deployed Qwen3-30B-A3B model~\cite{yang2025qwen3}. Following the template in \cref{fig:prompt1_supp}, the model generates a detailed textual description for the reference image. To align with e-commerce requirements, these prompts are specifically engineered to convey marketing intent and include rich details covering the subject, product, action, and lighting. This detailed prompt is then used with Qwen-Image~\cite{wu2025qwenimage} to synthesize the final reference image.

\textbf{Stage 2: Keyframe Synthesis.} With the reference image and original product information, we proceed to generate subsequent keyframes. We employ the commercial model Flux.1-Kontext~\cite{batifol2025kontext} for this task, chosen for its strong capability to maintain strict consistency with a reference image. The prompt generation for this stage is a two-step process:

Scenario Recommendation: First, using the product information and the reference image, we query our Qwen3-30B-A3B model with the template shown in \cref{fig:prompt2_supp}. The model suggests a series of scenes that are contextually appropriate for the product.

Editing Instruction Generation: Next, these recommended scenarios are transformed into precise editing instructions for Flux.1-Kontext using the template in \cref{fig:prompt3_supp}. The specific formatting in this template is crucial for ensuring the instructions are correctly interpreted by the generation model. Notably, we include the string \texttt{IMG\_1018.CR2} in the prompt, a technique we empirically found to enhance output image quality.

Through extensive experimentation, we summarize several critical insights:
\begin{itemize}
    \item \textbf{Isolated Scenario Planning:} Separating scenario recommendation as an independent subtask proves essential, as inappropriate settings lead to unrealistic outputs resembling mere copy-paste effects.

    \item \textbf{Reference Image Criteria:} To ensure successful subsequent editing, reference images should preferably feature full-body frontal poses of single subjects, avoiding multiple entities or reflective surfaces.

    \item \textbf{Instruction-Oriented Keyframe Prompts:} Keyframe generation benefits from imperative-style descriptions that explicitly guide content creation.

    \item \textbf{Consistency Preservation:} Maintaining cross-frame consistency requires avoiding detailed character and apparel specifications, instead emphasizing scene interactions and background elements.

    \item \textbf{Input Sensitivity of the Generation Model:} We observed that Flux.1-Kontext is highly sensitive to the reference image. For optimal results, the input must be a single-subject, frontal-view photograph with a natural expression. Reference images containing multiple individuals or unconventional poses consistently lead to generation failures.

    \item \textbf{Subject-Agnostic Prompting for Consistency:} To preserve the subject's facial identity and apparel, it is crucial to avoid describing these attributes in the keyframe prompts. Instead, we refer to the subject generically (e.g., starting the prompt with "This person...") and focus exclusively on describing the new scene, action, or camera angle. This prevents the model from regenerating the subject's appearance, thereby ensuring cross-frame consistency.
\end{itemize}

\textbf{Data Filtering.} Despite our meticulous prompt engineering, a subset of generated images may still fail to meet the stringent quality and consistency standards required for e-commerce. Consequently, we introduce a final data filtering stage. This process addresses two main issues: (1) common-sense violations, such as a subject wearing short sleeves in a snowy winter scene, and (2) common generative artifacts, particularly anatomical inconsistencies like unnatural limbs. This automated filtering is carried out by the Qwen2.5-VL-32B~\cite{bai2025qwen25vl} model to ensure the final dataset's high quality.

% \textbf{Dataset Visualization.} To demonstrate the quality and diversity of our constructed data, we conclude this section with a visual showcase of our final dataset. The displayed examples cover a wide array of product categories, including men's, women's, and children's apparel, loungewear, footwear, and accessories (\cref{fig:dataset}).

% \noindent\textbf{Why EAVSD?}
\textbf{Why EAVSD?} 
Existing datasets are not well suited for long-range narrative generation. 
\textit{CoMM} suffers from limited visual quality, 
\textit{CI-VID} contains only short sequences (typically fewer than three shots per instance), 
and \textit{OmniGen2} exhibits substantial textual redundancy that limits narrative diversity. 
These limitations restrict their applicability to professional long-range visual storytelling tasks.

In contrast, EAVSD is specifically designed to support such scenarios. 
It provides (1) \textbf{high-quality} visuals with rich professional annotations, 
(2) \textbf{long-range} sequences with an average of more than eight shots per instance, and 
(3) structured \textbf{narrative logic} aligned with professional production workflows.

As detailed in this section, we have already demonstrated the quality of our data construction pipeline, including model selection, prompt design, and filtering strategies. 
% To further substantiate the advantages of EAVSD, we present additional quantitative comparisons below.

\begin{figure*}[!ht]
  \centering
  \vspace{-6mm}
    \includegraphics[width=.7\linewidth]{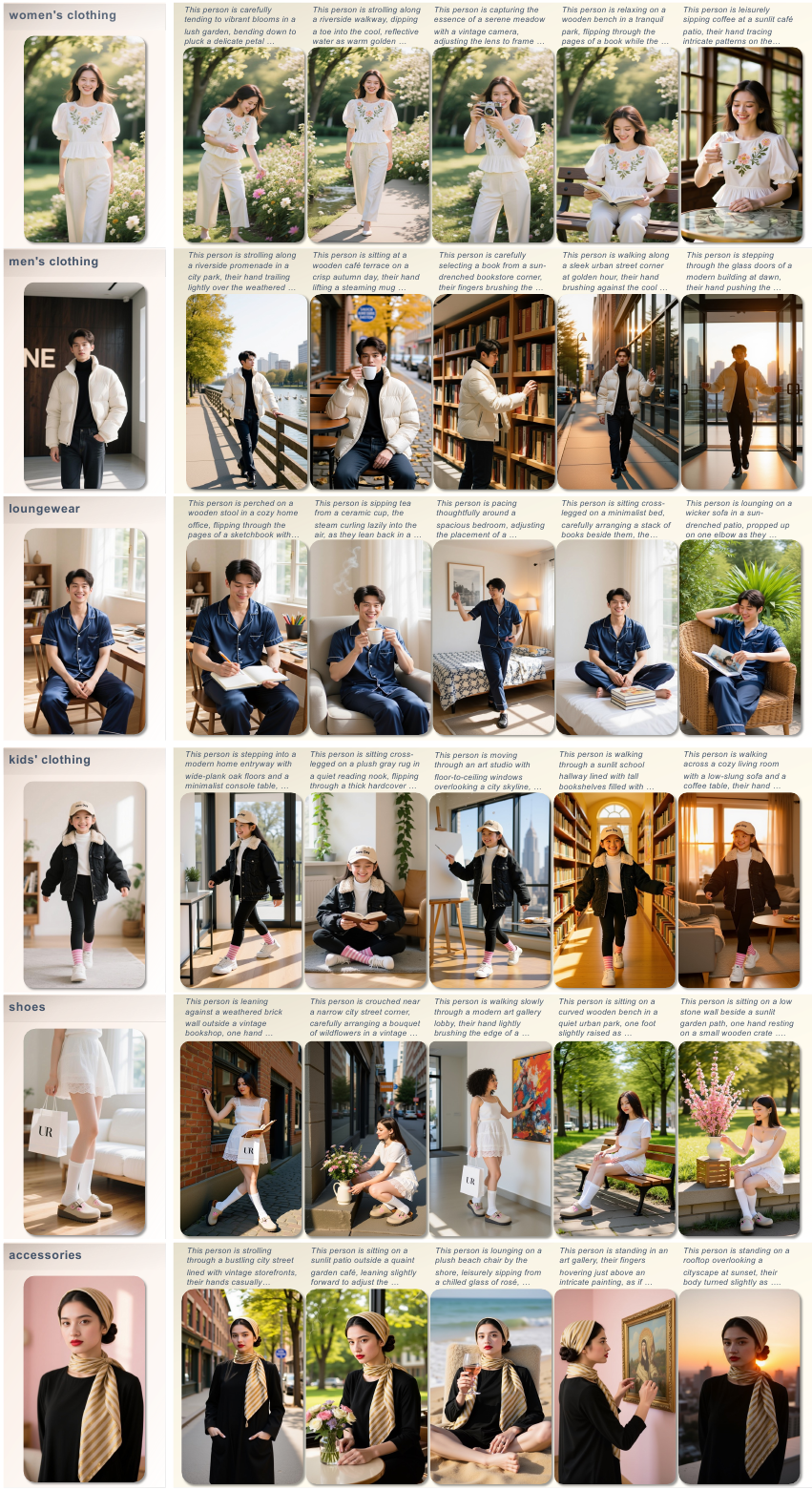}
    \vspace{-3mm}
    \caption{
        \textbf{Sample sequences from our EAVSD dataset.}
        The figure showcases the dataset's diversity across multiple e-commerce categories. Each row displays a sequence where a consistent subject and product are placed in various scenes, guided by descriptive text prompts. The dataset is designed to train models on tasks requiring high visual consistency while allowing for controlled narrative changes in action and setting, which is critical for advertising applications.
    }

  \label{fig:dataset}
\end{figure*}

\textbf{Dataset Visualization.} 
Our constructed dataset currently comprises 36K samples, totaling approximately 330K high-quality images, and we plan to expand it with additional product categories in the future. To demonstrate its quality and diversity, we conclude this section with a visual showcase of our final dataset (\cref{fig:dataset}). The displayed examples cover a wide array of the existing categories, including men's, women's, and children's apparel, loungewear, footwear, and accessories.

\clearpage
\clearpage

\newpage
\section{Evaluation Details}
\label{sec:eva_details}
This section provides further details on our dataset processing and evaluation metrics for the experiment in \cref{sec:experiments}.

\subsection{Consistent Visual Generation (Q1)}
\textbf{Test Set Curation.} We observed that the original OmniGen2~\cite{wu2025omnigen2} pre-training data contains a significant number of low-quality samples. To establish a more reliable benchmark, we manually curated a test set of 100 sequences, each comprising alternating text prompts and video frames. Our curation process specifically prioritized samples that demand strong inter-frame consistency, thereby enabling a focused evaluation of Narrative Weaver's capabilities.

\begin{figure}
  \centering
    \includegraphics[width=1.0\linewidth]{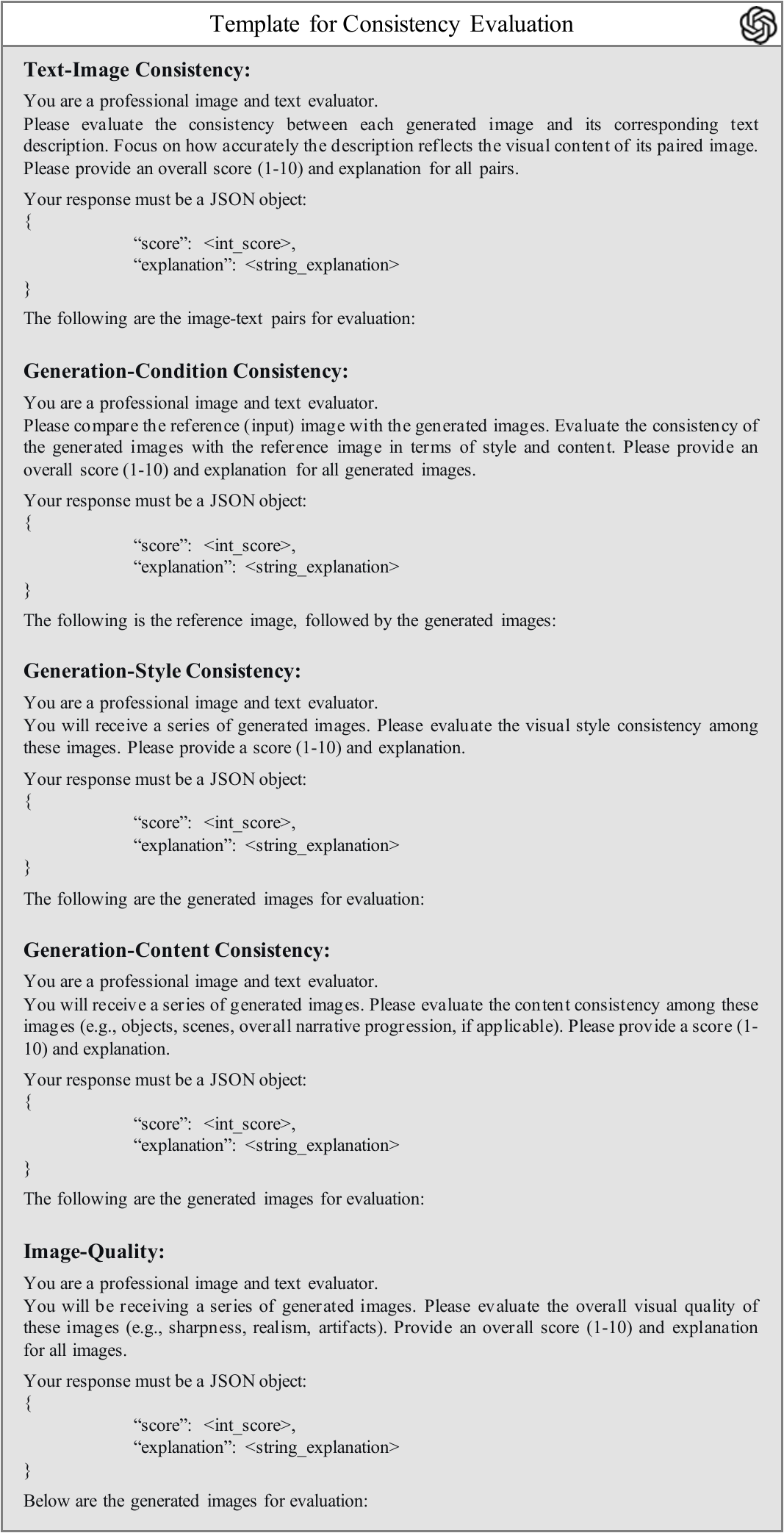}
  \caption{The prompt template provided to GPT-4o for our consistency evaluation. It requires the model to score instruction following, consistency, and image quality, and to provide a rationale for each score.}
  \label{fig:task1_prompt}
\end{figure}

\textbf{LLM-based Evaluation.} We provide the prompt template used for our LLM-based evaluation in \cref{fig:task1_prompt}. This prompt is meticulously designed to comprehensively assess the model's performance across three key dimensions: instruction following, consistency preservation, and image quality. To ensure the validity and reliability of the automated scoring, we instruct the language model to provide a detailed rationale for each assigned score. This practice significantly enhances the stability and trustworthiness of the evaluation process.

\textbf{Baseline Implementation Details.} For a fair comparison, we reproduced several baseline methods. In the following, we describe their implementation details.
\begin{itemize}
    \item \textbf{StoryDiffusion~\cite{zhou2024storydiffusion}:} This method first generates an initial image from the text prompt corresponding to the reference image. It then utilizes the intermediate tokens from this initial generation process to condition the creation of all subsequent images.

    \item \textbf{AnimeShooter~\cite{qiu2025animeshooter}:} The original implementation of AnimeShooter trains a specific LoRA module for each film or IP to achieve high fidelity. To evaluate its generalization capabilities in a broader context, we omitted this LoRA module in our experiments.

    \item \textbf{Reference-based Methods (IP-Adapter~\cite{ye2023ip-adapter}, Flux.1-Kontext~\cite{batifol2025kontext}, Qwen-Image-Edit~\cite{wu2025qwenimage}):} This category of methods, including IP-Adapter, Flux.1-Kontext, and Qwen-Image-Edit, conditions the generation of each new image on both the initial reference image and the current text prompt. While this approach effectively preserves consistency between each generated image and the initial reference, it often struggles to maintain consistency among the generated images themselves.
\end{itemize}
Unless a model was specifically trained at a fixed resolution, all baselines were configured to generate images at the same resolution as the provided condition image.

\textbf{User Study Details.}
For our human evaluation, we compared Narrative Weaver with the three best-performing methods from Q1 (Flux.1-Kontext~\cite{batifol2025kontext}, Qwen-Image-Edit~\cite{wu2025qwenimage}, and StoryDiffusion~\cite{zhou2024storydiffusion}). Each survey presented participants with the outputs from these four methods for a randomly selected test case. The order of the results was randomized to prevent bias. Participants were asked to choose the most preferable result overall. The final results were compiled from over 180 valid survey responses.

\subsection{Autonomous Narrative Planning (Q2)}
\textbf{CoMM Dataset Processing.} To evaluate Narrative Weaver's autonomous narrative generation capability (Q2 in \cref{sec:experiments}), we employ the CoMM dataset~\cite{chen2025comm}. The original dataset is compiled from diverse sources and suffers from significant data imbalance due to invalid URLs. We therefore selected two instruction-based subsets, Instructables and WikiHow, which are particularly well-suited for assessing the model's problem-solving and narrative planning capacity. For a fair comparison, we re-evaluated the official test set using the weights provided by the original benchmark authors.

During data processing, we standardized each sample by limiting it to a maximum of 16 images and the \texttt{"step\_info"} field to 12 elements. For continuation tasks, we generated training samples by randomly truncating text-image sequences, using the first half as input and the second half as the target output, ensuring each target contained at least one image. After filtering for invalid images, this procedure yielded approximately 170K training samples. For question-based response tasks, a similar filtering process resulted in approximately 150K training samples. In this experiment, all images were rescaled to a resolution of 512×512 pixels to maintain consistency with the original benchmark.

% 对于domain 外的数据EMU2完全做不了。现象是不生成文本，说明不具备规划能力，或者进行复读，或者完全不生成文本。另外模型不知道何时该输出图片，有可能一下子把所有文本步骤都生成出来
\textbf{A Note on the EMU2 Baseline.} The official CoMM benchmark includes EMU2~\cite{sun2024emu2}, a 33B large-scale unified model. However, we excluded it from our comparison for two primary reasons. First, the official repository does not provide the specific checkpoint or training code used for the benchmark, hindering reproducibility. Second, our preliminary tests with the publicly available EMU2 weights revealed significant failure modes: the model often failed to generate any text, produced repetitive content, or demonstrated a lack of planning ability by generating all text steps at once without interleaving images. Given these issues, reporting its scores would compromise the integrity of our evaluation.

\textbf{CI-VID Dataset.} For video narrative generation, we utilized the CI-VID dataset~\cite{ju2025ci-vid}, which contains video clips with corresponding captions and inter-clip transition descriptions. To construct our training samples and mitigate potential black screen issues, we consistently selected the fifth frame from each video segment. The textual input for the initial frame is its corresponding clip's caption, while the guidance for all subsequent frames comes from the transition descriptions between clips. The first frame serves as the condition for generating the rest of the sequence. All training data from this dataset used a 480p anchor resolution (480×854), with the original video aspect ratio preserved.

\subsection{Extended Application Scenarios (Q3)}
To demonstrate the practical utility of our method, we apply Narrative Weaver to the domain of e-commerce advertising. By leveraging its dual capabilities in autonomous narrative planning and controllable consistency generation, our objective is to produce sequences of visual content that can serve as keyframes for complete advertising videos.

\textbf{Evaluation Setup.} For this application, we employ the same evaluation metrics as those used for Q1 in \cref{sec:experiments}. We constructed a dedicated test set by randomly sampling 200 sequences from our generated e-commerce data.

\textbf{Qualitative Comparison.} While the main paper presented only a limited number of examples due to space constraints, this appendix provides a more comprehensive qualitative comparison. \cref{fig:task3_compare} showcases a side-by-side comparison between the results generated by Narrative Weaver and those from a leading image editing model. Since Qwen-Image-Edit lacks autonomous text generation capabilities, we supplied it with the same prompts generated by Narrative Weaver to ensure a fair comparison.

\section{Experimental Details}
\label{sec:experimental_details}
Our multi-stage training strategy is designed to decouple Narrative Planning (Stage 1) from Visual Generation (Stages 2 and 3). This separation is enabled by a carefully designed attention mask that effectively freezes the narrative planning capability of the language model after Stage 1. Consequently, the subsequent stages can focus exclusively on enhancing coherent visual content generation without compromising the already-learned textual planning abilities.

We illustrate our training recipe using the e-commerce dataset (Q3) as a representative example. The detailed hyperparameters for each stage are provided in \cref{tab:training_recipe}.
\begin{table}[!ht]
    \caption{Training recipe of Narrative Weaver.}
    \label{tab:training_recipe}

    \centering
    \resizebox{0.45\textwidth}{!}{
    \begin{tabular}{l|cccc}
        \toprule
         \textbf{Hyperparameters} & Stage-1 & Stage-2.1 & Stage-2.2 & Stage-3  \\
         \midrule
         Learning rate & $1\times 10^{-5}$ & $5\times 10^{-5}$ & $1\times 10^{-5}$ & $1\times 10^{-6}$ \\ 
         LR scheduler & Constant & Cosine & Constant & Constant\\
         Weight decay & 0.0 & 0.0 & 0.0 & 0.0 \\
         Gradient norm clip & 1.0  & 1.0  & 1.0  & 1.0 \\
         Optimizer & \multicolumn{4}{c}{AdamW ($\beta_1 = 0.9, \beta_2 = 0.99, \epsilon = 1 \times 10^{-8}$)} \\
         Loss type & CE & MSE & MSE & MSE \\
         Warm-up steps & 0 & 0 & 100 & 100 \\
         Training steps & 32K & 400K & 48K & 32K \\
         Batch size & 8 & 128 & 8 & 8 \\
         Module & Qwen2.5VL-3B & \multicolumn{2}{c}{Learnable Query} & Flux.1-Dev\\
         
        \bottomrule
    \end{tabular}
    }
\end{table}

\textbf{Stage 1:} Narrative Planning. In this stage, we train the Large Vision-Language Model (Qwen2.5VL-3B) on the task-specific dataset to master narrative and textual planning.

\textbf{Stage 2:} Bridging Language and Vision. This stage connects the planner with the visual generator and is divided into two sub-stages:
Stage 2.1 (Connector Pre-training): We first pre-train the Learnable Queries on a large-scale public image-text dataset. The objective is to align these queries with the text embedding space of the visual generation model (Flux.1-Dev). Crucially, this pre-training is a one-time process. The resulting Learnable Queries can be seamlessly reused as a plug-and-play module for various downstream fine-tuning tasks, eliminating the need for repeated training.
Stage 2.2 (Task-specific Fine-tuning): The pre-trained Learnable Queries are then fine-tuned on the small, task-specific e-commerce dataset to adapt them to the specific domain.

% \begin{table*}[!t]
%     \caption{Efficiency.}
%     \label{tab:training_recipe}

%     \centering
%     \resizebox{0.85\textwidth}{!}{
%     \begin{tabular}{l|c|ccccccccccccccc}
%         \toprule
%          Implementation &  Keyframe Num & 1 & 2 & 3 & 4 & 5 & 6 & 7 & 8 & 9 & 10 & 11 & 12 & 16  \\
%          \midrule
%          \multirow{2}{*}{\textbf{Vanilla}} & TFLOPs &  82 & 230 & 450 & 744 & 1112 & 1553 & 2068 & 2656 & 3318 & 4053 & 4862 & 5744 \\ 
%          & TMACs & 41 & 115 & 225 & 372 & 556 & 776 & 1034 & 1328 & 1659 & 2026 & 2431 & 2872 \\
%          \cmidrule(lr){1-1}\cmidrule(lr){2-2} \cmidrule(lr){3-14}
%          \multirow{2}{*}{\textbf{Ours}} & TFLOPs& & &248 &331 &441 &497 &580 &663\\
%          & TMACs &   & &124 &165 &207 &248 &290 &331 \\
%         \bottomrule
%     \end{tabular}
%     }
% \end{table*}

\textbf{Stage 3:} Visual Generation Fine-tuning. Finally, we fine-tune the visual generation model (Flux.1-Dev) itself, further adapting it to the domain while keeping the other modules frozen.

This modular design makes the overall training process highly efficient for adapting to new tasks, as the Learnable Query pre-trained with large-scale data in Stage 2.1 can be seamlessly reused across diverse tasks without the need for repeated training. For all experiments presented in the main paper, we adopted a consistent image processing protocol. We used an anchor resolution of 480p (480×854), resizing all frames while preserving their original aspect ratio.

\section{Additional Experimental Results}
\label{sec:Add_results}

In this section, we provide extensive qualitative results to further substantiate the findings presented in the main paper. We offer more visualizations for each of our core experimental setups: controllable consistent generation (Q1), autonomous narrative generation (Q2), and the e-commerce application (Q3).

\begin{figure*}[!ht]
  \centering
    \includegraphics[width=.95\linewidth]{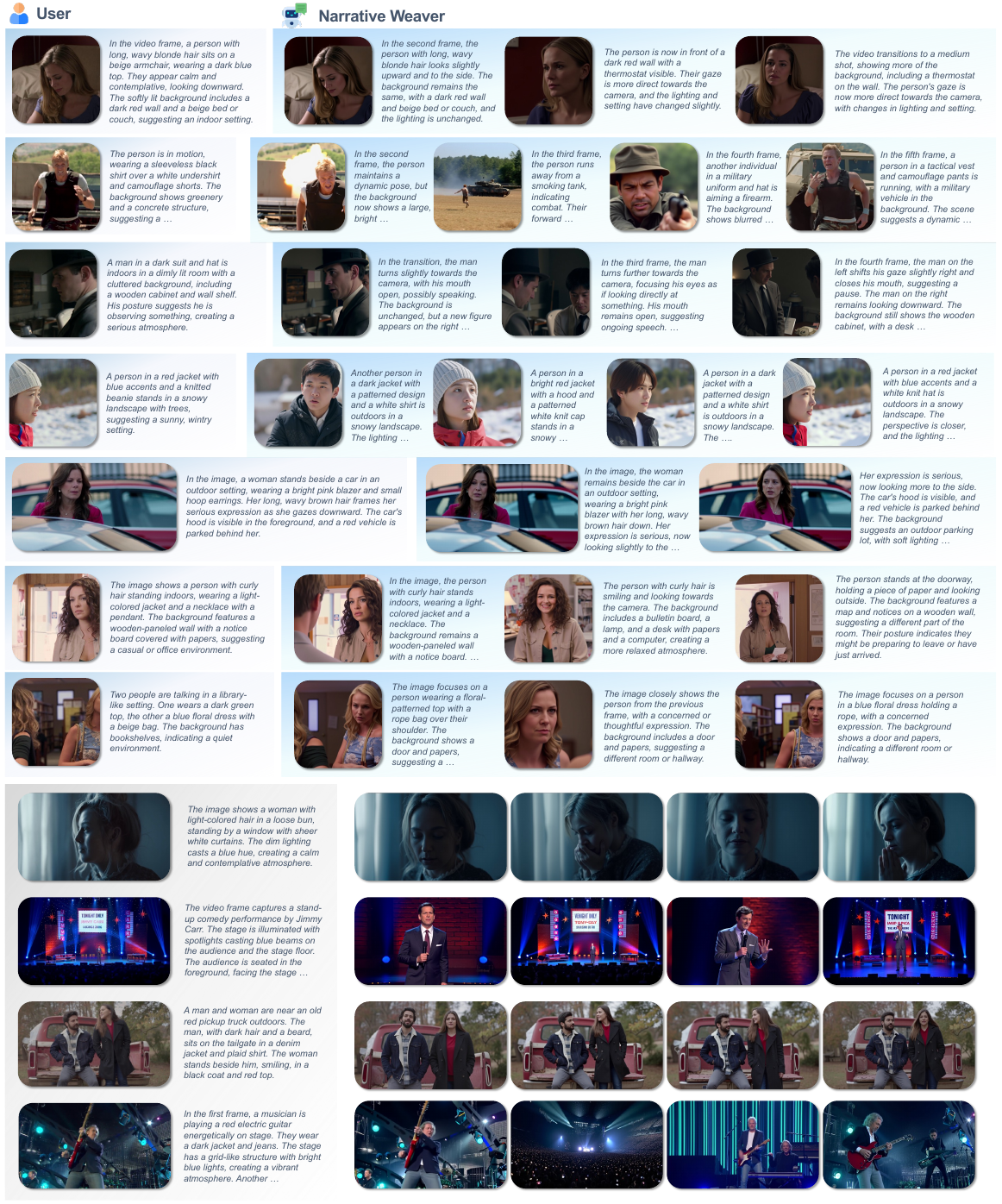}
    % \vspace{-2mm}
  \caption{
  \textbf{Additional qualitative results of multi-frame narrative generation by Narrative Weaver.}
    Each row showcases a complete generation sequence. The leftmost column presents the user's initial input (a reference image and its description). The subsequent columns display the multi-frame visual narrative autonomously generated by our model, including both the synthesized images and their corresponding textual descriptions. These diverse examples highlight Narrative Weaver's proficiency in maintaining high cross-frame consistency—preserving subject identity, apparel, and key background elements—while coherently evolving the narrative through subtle changes in pose, expression, and camera perspective.
  }
  \label{fig:task1_more}
\end{figure*}

\begin{figure*}[!ht]
  \centering
    \includegraphics[width=.72\linewidth]{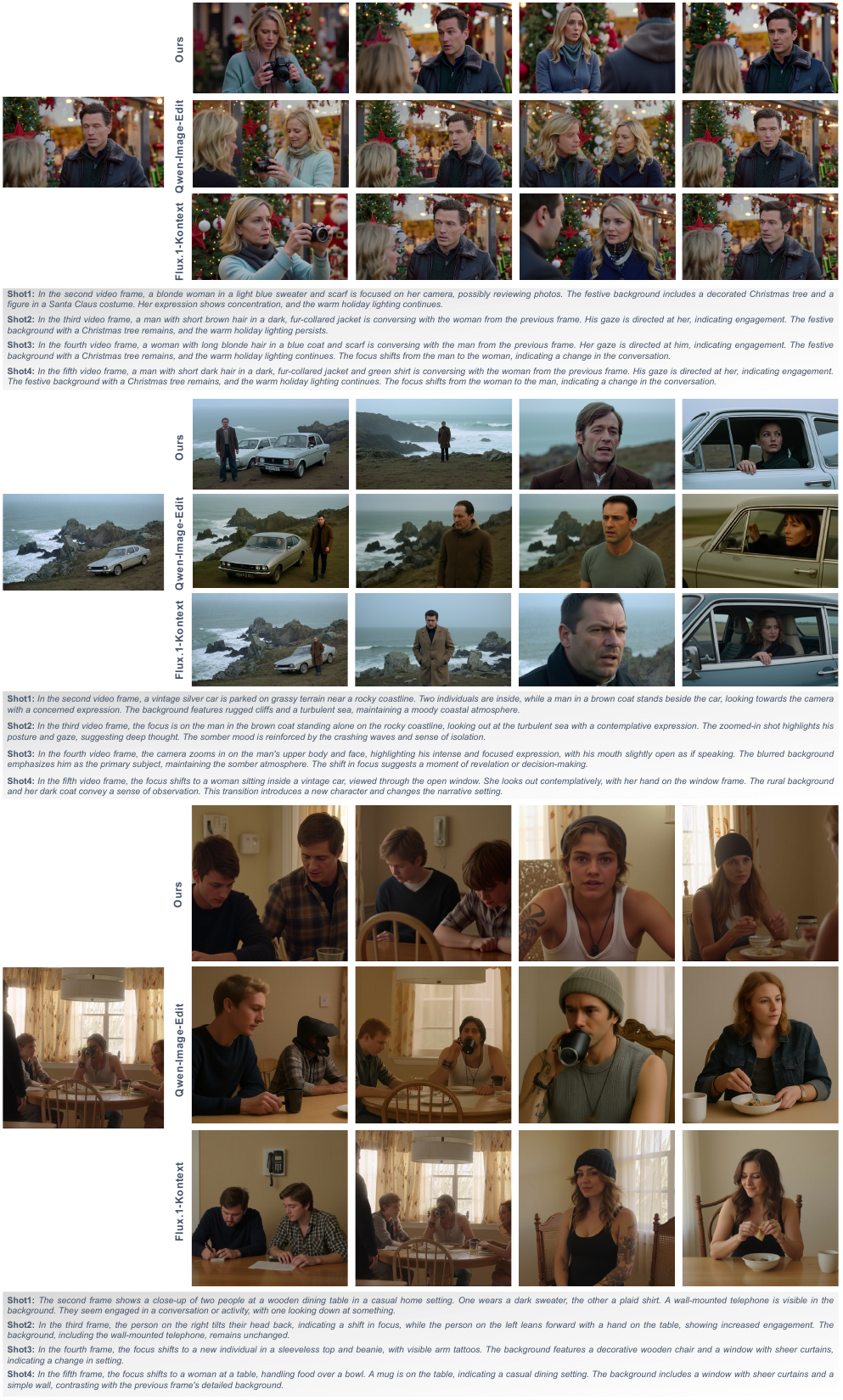}
    \vspace{-2mm}
\caption{
    \textbf{Comparison with leading image editing models on multi-frame narrative generation.}
    While all models exhibit strong adherence to the text prompts and maintain high subject consistency, Narrative Weaver uniquely generates outputs with a more \textbf{cinematic and aesthetically pleasing quality}. Note our model's superior handling of lighting, color, and composition, which contributes to a more authentic, film-like visual narrative compared to the often more literal or digitally rendered feel of the baseline results.
}
    \label{fig:task1_compare}
\end{figure*}

\textbf{For controllable consistent generation (Q1)}, we first present an expanded set of qualitative results from Narrative Weaver in \cref{fig:task1_more}. These diverse examples further demonstrate the model's proficiency in maintaining high cross-frame consistency in subject identity, apparel, and background, while coherently evolving the narrative according to user prompts. Following this, \cref{fig:task1_compare} provides a side-by-side qualitative comparison with leading baseline models. This visualization highlights that while competitors can adhere to prompts, Narrative Weaver uniquely achieves a more cinematic and aesthetically pleasing quality in its outputs, showcasing superior handling of lighting, color, and composition.

\begin{figure*}[!ht]
  \centering
    \includegraphics[width=.95\linewidth]{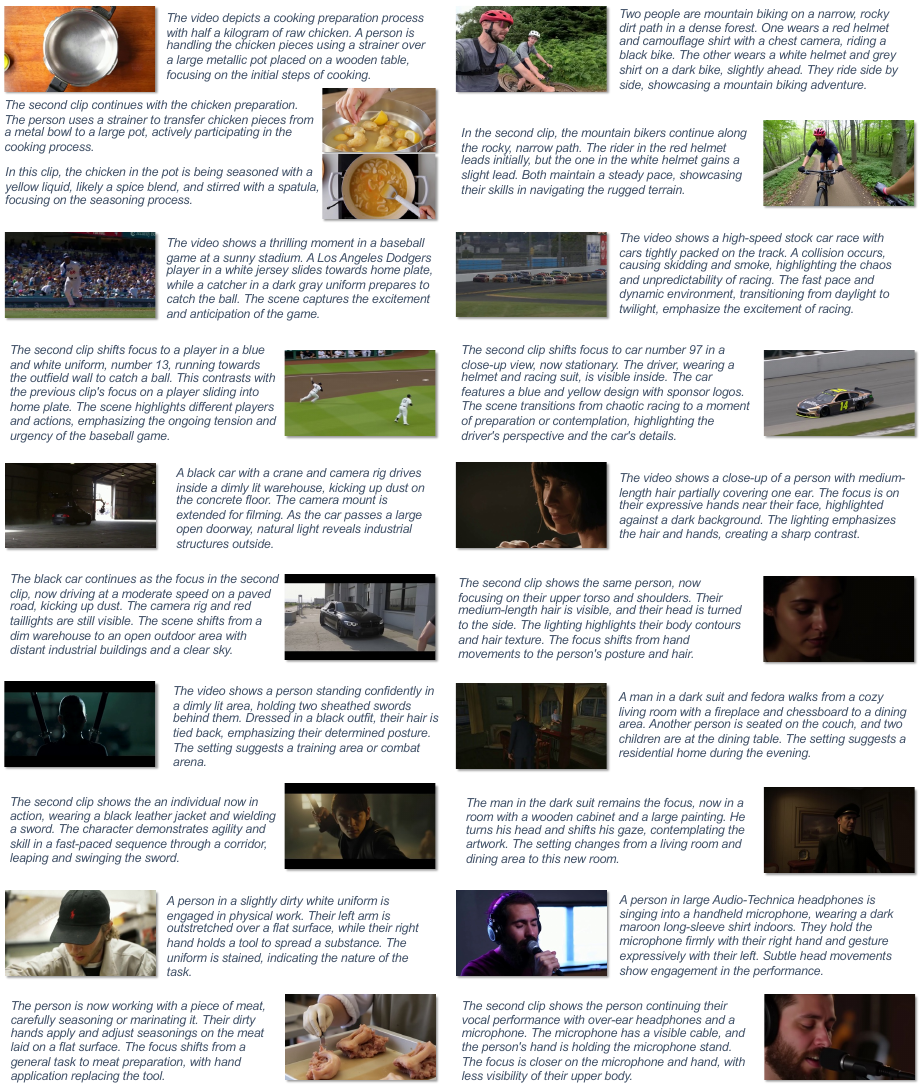}
  \caption{
        \textbf{Qualitative examples of autonomous story continuation by Narrative Weaver.}
        Given only the first frame and text as input for each sequence, Narrative Weaver autonomously plans and generates a coherent multi-frame continuation. The diverse examples—from procedural tasks like cooking to dynamic events like sports—showcase the model's robust planning capabilities. Its ability to logically advance a narrative by continuing actions, introducing new elements, or shifting focus highlights its understanding of storytelling beyond simple image editing.
    }
  \label{fig:task2_more}
\end{figure*}

\textbf{For the task of autonomous narrative generation (Q2)}, additional examples are showcased in \cref{fig:task2_more}. These results underscore the model's robust planning capabilities, showing its ability to logically and creatively continue a story from a single initial prompt across a variety of scenarios.

\begin{figure*}[!ht]
  \centering
    \includegraphics[width=.98\linewidth]{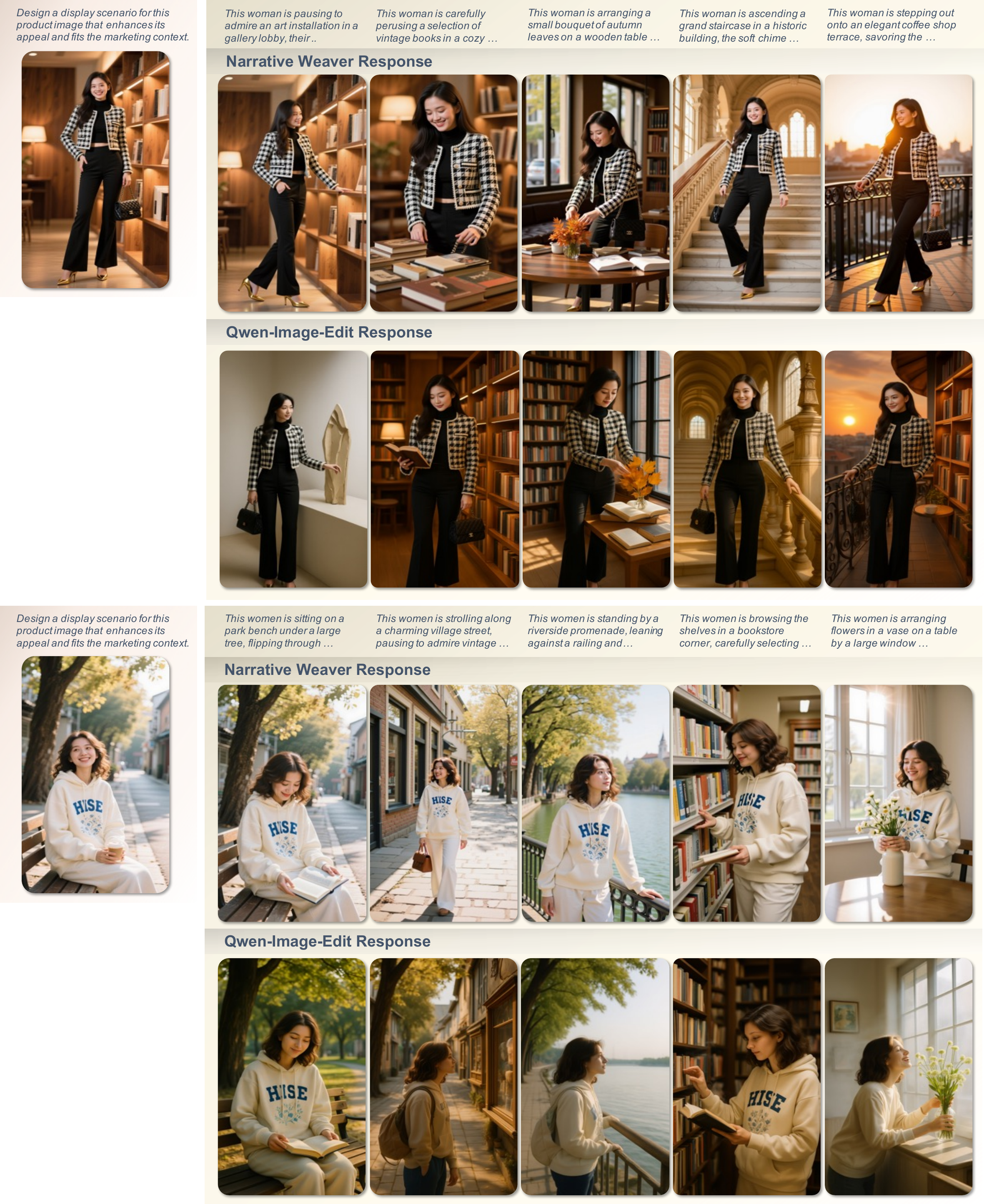}
    \vspace{-2mm}
  % \caption{\textbf{Comparison with a leading editing model, Qwen-Image-Edit.} Narrative Weaver not only demonstrates superior stylistic consistency with the conditional image but also excels in preserving key semantic details required by the task.}
  \caption{
    \textbf{Comparison with a leading editing model, Qwen-Image-Edit.}
    Narrative Weaver not only demonstrates superior stylistic consistency with the conditional image but also excels in preserving key semantic details required by the task. In contrast, Qwen-Image-Edit exhibits noticeable failure modes: it struggles with inconsistent color tones between frames (upper example) and introduces a warm color cast that deviates from the style of the reference image (lower example).
}
  \label{fig:task3_compare}
\end{figure*}

\textbf{Finally, regarding our e-commerce application (Q3)}, \cref{fig:task3_compare} presents a direct comparison against a leading editing model, Qwen-Image-Edit. This comparison illustrates our model's superior ability to preserve not only stylistic consistency with the reference image but also key semantic details required by the task, which is critical for real-world applications.

\textbf{Showcase of a Full Advertising Production Pipeline.}
To demonstrate the practical utility of Narrative Weaver in a real-world production workflow, we produced complete, ready-for-deployment advertising videos, which are available in the supplementary materials. Our end-to-end pipeline is as follows:

First, we leverage Narrative Weaver to generate the core visual content: a sequence of high-quality, consistent keyframes that define the narrative. Next, we employ a Large Language Model (LLM) to create coherent and contextually appropriate shot descriptions or transition narratives for these keyframes. These image-text pairs are then fed into the Wan2.2 model for video synthesis, which generates a short video clip for each keyframe. Finally, all resulting video clips are concatenated to form a seamless, complete advertising video.
% \begin{table}[!ht]
%     \caption{Evaluation of Text-Image Alignment in Consistent Keyframe Generation (Q1).}
%     \label{tab:task1_clip-t}
%     \vspace{-2mm}
%     \centering
%     \resizebox{0.45\textwidth}{!}{
%     \begin{tabular}{l|ccccc}
%         \toprule
%           &  \multicolumn{5}{c}{\textbf{CLIP Score - T} $\uparrow$} \\
%          Methods & shot-1 & shot-2 & shot-3 & shot-4 & Mean \\
%          \midrule
%          TALC~\cite{bansal2024talc}  & 21.99 & 22.12 & 21.22 & 22.14 & 22.09 \\ 
%          StoryDiffusion~\cite{zhou2024storydiffusion}  & 32.93 & 32.48 & 32.84 & 33.03 & 32.69 \\
%          IP-Adapter~\cite{ye2023ip-adapter}  & 31.06 & 31.34 & 31.19 & 30.70 & 31.10\\
%          AnimeShooter~\cite{qiu2025animeshooter}  & 24.91 & 25.25 & 24.81 & 24.19 & 24.90 \\
%          Flux.1-kontext~\cite{batifol2025kontext}  & 30.95 & 29.53 & 30.56 & 30.33 & 30.23 \\
%          Qwen-Image-Edit~\cite{wu2025qwenimage}  & 30.36 & 29.68 & 29.99 & 29.72 & 29.88 \\
%          \textbf{Narrative Weaver (Ours)}  & 30.88 & 29.86 & 30.68 & 30.34 & 30.35\\
%         \bottomrule
%     \end{tabular}
%     }
% \end{table}

\section{Detailed Ablation Results}
\label{sec:detail_ablation}

\begin{figure*}[!ht]
  \centering
    \includegraphics[width=.98\linewidth]{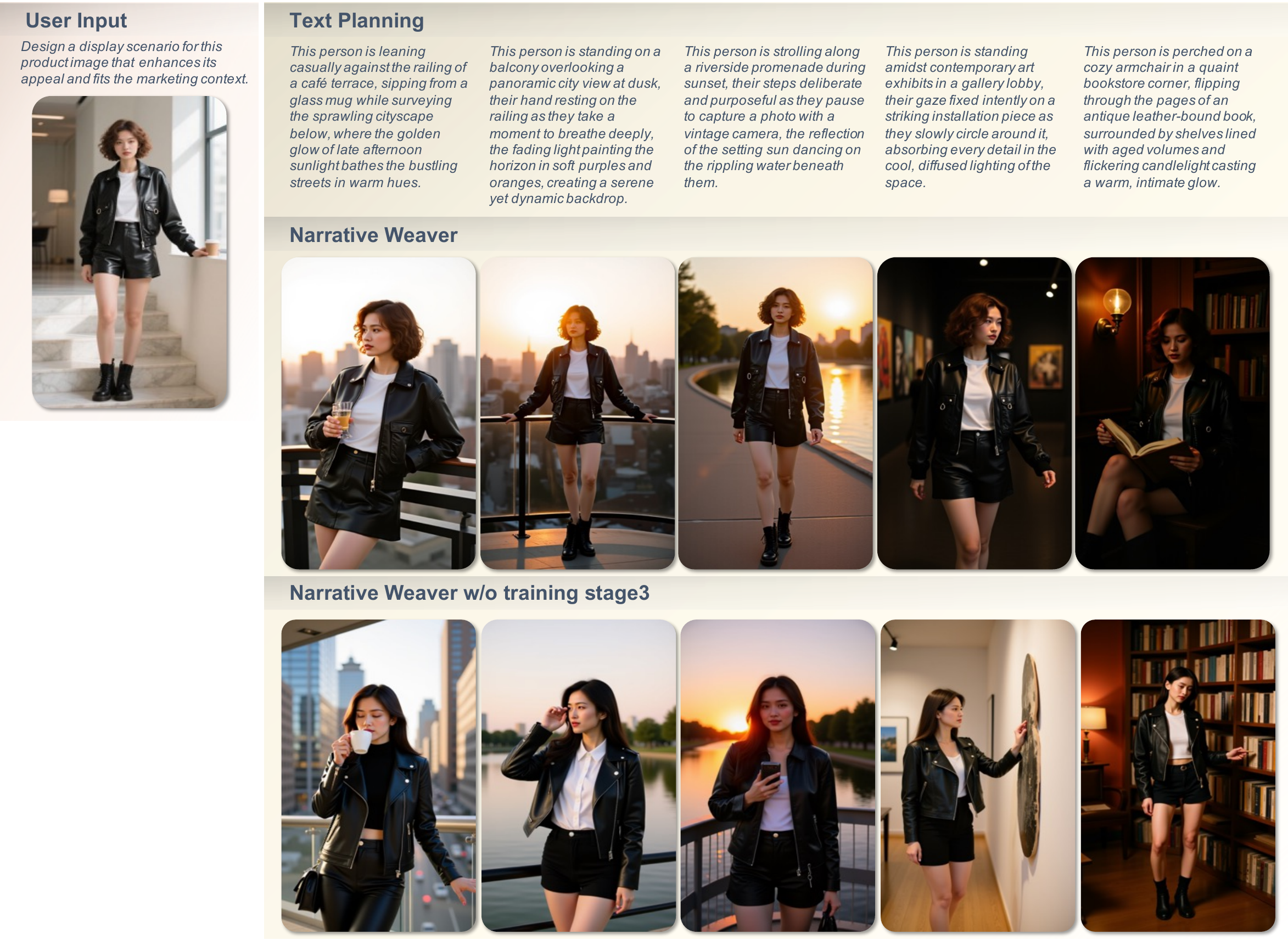}
    \vspace{-2mm}
    \caption{
    \textbf{Ablation Study Results.}
    Comparison between Narrative Weaver and its variant without Stage 3 training demonstrates that: (1) the first two stages establish fundamental semantic alignment; (2) Stage 3 significantly enhances inter-frame consistency; and (3) Stage 3 is crucial for imparting fine-grained control, as the variant without it produces images that deviate significantly from the reference.
    }

  % \caption{\textbf{Ablation Study Results.} Comparison between Narrative Weaver and its variant without Stage 3 training demonstrates that: (1) the first 2 stages establish fundamental semantic alignment capability, and (2) Stage 3 training significantly enhances inter-frame consistency.}
  \label{fig:ablation}
\end{figure*}

To isolate and verify the contribution of Stage 3 in our training pipeline, we present a direct comparison between the full Narrative Weaver model and a variant trained without this final stage. As shown in \cref{fig:ablation}, the model without Stage 3 can follow the core semantic instructions but fails to maintain strict visual consistency, leading to variations in subject appearance and details across frames.

The inclusion of Stage 3 rectifies this issue by enabling fine-grained control over the visual generation process. This results in a dramatic enhancement in the model's ability to preserve inter-frame consistency, which is critical for creating coherent visual narratives. This ablation clearly demonstrates that Stage 3 is essential for achieving the high-fidelity consistency that is a core strength of our method.

\clearpage

\begin{table*}[!t]
    % \caption{Efficiency.}
    \caption{
    \textbf{Computational cost (TFLOPs) as a function of the number of generated keyframes.}
    Our approach demonstrates significantly better scalability compared to a vanilla self-attention baseline. The computational cost of our method grows linearly with the sequence length, unlike the vanilla implementation, whose cost grows quadratically, making our method far more efficient for longer sequences.
}
    \label{tab:efficiency_comparison}
    \vspace{-2mm}
    \centering
    \resizebox{0.9\textwidth}{!}{
    \begin{tabular}{l|c|cccccccccccccccc}
        \toprule
         Implementation &  Keyframe Num & 1 & 2 & 3 & 4 & 5 & 6 & 7 & 8 & 9 & 10 & 11 & 12 & 16 & 20 \\
         \midrule
         \textbf{Vanilla} & \multirow{2}{*}{TFLOPs} &  82 & 230 & 450 & 744 & 1112 & 1553 & 2068 & 2656 & 3318 & 4053 & 4862 & 5744 &10008 & 15449\\ 
         \cmidrule(lr){1-1} \cmidrule(lr){3-16}
         \textbf{Ours} &  &82 &165 &248 &331 &441 &497 &580 &663 & 746 & 829 & 912 & 995 & 1077 & 1160\\
        \bottomrule
    \end{tabular}
    }
\end{table*}

\clearpage

\section{Additional Efficiency Analysis}
\label{sec:efficiency_analysis}
% 为了客观的比较，可以与ic-lora进行对比
The superior efficiency of Narrative Weaver, as demonstrated in \cref{tab:efficiency_comparison}, stems from a fundamental architectural choice. Specifically, our approach delegates the task of establishing cross-frame coherence to the Multimodal Large Language Model (MLLM) planning stage. As a result, the input token sequence for the Diffusion Transformer (DiT) remains constant in length, regardless of the number of keyframes being generated.

In contrast, vanilla implementations~\cite{huang2024gdt,huang2024iclora} must maintain coherence within the DiT itself. This is typically achieved by concatenating the latent representations of all preceding frames and processing them simultaneously, causing the sequence length to grow with each new frame. This architectural difference directly leads to the quadratic explosion in computational complexity observed for the vanilla method in \cref{tab:efficiency_comparison}, whereas our approach maintains a highly efficient, near-linear scaling.

\section{Limitations and Future Works}
\label{sec:limitaions}
% 错误案例表明模型能力边界
% 关键帧生成到视频生成之间的gap

In this paper, we introduced Narrative Weaver, a unified framework capable of fine-grained control, long-range consistency preservation, and autonomous narrative planning. While the architecture is theoretically capable of generating any form of visual content, our current work presents a preliminary implementation focused exclusively on images.

We acknowledge that focusing on long-range \textit{image} consistency is a pragmatic choice, largely dictated by current resource constraints. A critical and promising direction for future work is to extend Narrative Weaver to ensure consistency across multiple \textit{video clips}. This extension is vital because video introduces crucial elements of temporal consistency, including coherent character motion and logical camera movements (cinematography), which are not captured in static images. We leave this ambitious extension for future investigation.

Furthermore, our research highlights a significant challenge in the field: the scarcity of high-quality datasets designed for controllable, long-range consistent content generation. To address this gap, we constructed a new dataset tailored to the e-commerce domain. However, the development of similar large-scale, diverse datasets for broader application domains remains a critical need for advancing research in this area and represents another important avenue for future work.

% \section{Rationale}
% \label{sec:rationale}
% % 
% Having the supplementary compiled together with the main paper means that:
% % 
% \begin{itemize}
% \item The supplementary can back-reference sections of the main paper, for example, we can refer to \cref{sec:intro};
% \item The main paper can forward reference sub-sections within the supplementary explicitly (e.g. referring to a particular experiment); 
% \item When submitted to arXiv, the supplementary will already included at the end of the paper.
% \end{itemize}
% % 
% To split the supplementary pages from the main paper, you can use \href{https://support.apple.com/en-ca/guide/preview/prvw11793/mac#:~:text=Delete%20a%20page%20from%20a,or%20choose%20Edit%20%3E%20Delete).}{Preview (on macOS)}, \href{https://www.adobe.com/acrobat/how-to/delete-pages-from-pdf.html#:~:text=Choose%20%E2%80%9CTools%E2%80%9D%20%3E%20%E2%80%9COrganize,or%20pages%20from%20the%20file.}{Adobe Acrobat} (on all OSs), as well as \href{https://superuser.com/questions/517986/is-it-possible-to-delete-some-pages-of-a-pdf-document}{command line tools}.

\end{document}